%% file: main.tex
\documentclass[runningheads]{llncs}

\usepackage{eccv}

\usepackage{eccvabbrv}

\usepackage{graphicx}
\usepackage{booktabs}

\usepackage[accsupp]{axessibility}  % Improves PDF readability for those with disabilities.

\usepackage{hyperref}

\usepackage{orcidlink}

\usepackage{multirow}

\begin{document}

% ---------------------------------------------------------------
% TODO REVIEW: Replace with your title
\title{TPA3D: Triplane Attention for\\Fast Text-to-3D Generation} 

% TODO REVIEW: If the paper title is too long for the running head, you can set
% an abbreviated paper title here. If not, comment out.
% \titlerunning{Triplane Attention for Fast Text-to-3D Generation}

% TODO FINAL: Replace with your author list. 
% Include the authors' OCRID for the camera-ready version, if at all possible.
\author{Bin-Shih Wu\inst{1^*}\orcidlink{0009-0004-7557-7637}, 
Hong-En Chen\inst{1^*}\orcidlink{0000-0001-8787-4234},
Sheng-Yu Huang\inst{1}\orcidlink{0000-0002-3149-9620}, \\
and Yu-Chiang Frank Wang\inst{1,2}\orcidlink{0000-0002-2333-157X}}

% TODO FINAL: Replace with an abbreviated list of authors.
\authorrunning{B.~Wu et al.}
% First names are abbreviated in the running head.
% If there are more than two authors, 'et al.' is used.

% TODO FINAL: Replace with your institution list.
\institute{
National Taiwan University \and
NVIDIA \\
\email{\{r12942090, b08901058, f08942095\}@ntu.edu.tw} \\
\email{frankwang@nvidia.com}}

\maketitle
\def\thefootnote{*}\footnotetext{These authors contributed equally to this work.}

\input{sec/0_abstract}
\input{sec/1_introduction}

\input{sec/2_related_works}
\input{sec/3_preliminary}
\input{sec/4_methods}
\input{sec/5_experiments}

\input{sec/6_conclusion}

% ---- Bibliography ----
%
% BibTeX users should specify bibliography style 'splncs04'.
% References will then be sorted and formatted in the correct style.
%
% \input{main.bbl}
\bibliographystyle{splncs04}
\bibliography{main}

\input{sec/X_supp}

\end{document}

%% file: sec/0_abstract.tex
\begin{abstract}
Due to the lack of large-scale text-3D correspondence data, recent text-to-3D generation works mainly rely on utilizing 2D diffusion models for synthesizing 3D data. Since diffusion-based methods typically require significant optimization time for both training and inference, the use of GAN-based models would still be desirable for fast 3D generation. In this work, we propose Triplane Attention for text-guided 3D generation (TPA3D), an end-to-end trainable GAN-based deep learning model for fast text-to-3D generation. With only 3D shape data and their rendered 2D images observed during training, our TPA3D is designed to retrieve detailed visual descriptions for synthesizing the corresponding 3D mesh data. This is achieved by the proposed attention mechanisms on the extracted sentence and word-level text features. In our experiments, we show that TPA3D generates high-quality 3D textured shapes aligned with fine-grained descriptions, while impressive computation efficiency can be observed.
\keywords{3D computer vision \and text-to-3D generation}
\end{abstract}

%% file: sec/1_introduction.tex
\section{Introduction}
\label{sec:intro}

3D object generation has become a thriving area of computer vision research in recent years, particularly with the increasing prevalence and requirement of AR/VR technologies, video game development, movie visual effects, and robotic simulations~\cite{katara2023gen2sim, ha2021fit2form}. In the pursuit of automating the creation of 3D objects, numerous researchers strive to develop approaches for generating high-quality 3D assets. Early methods in 3D object generation~\cite{chan2022efficient, gao2022get3d, mescheder2019occupancy, mildenhall2021nerf, chen2019learning, achlioptas2018learning, pavllo2021learning, yang2019pointflow} predominantly focus on learning representations that are both efficient and effective for generating 3D objects. However, the inherent unconditionality of these methods not only impedes the customization of generated shapes based on specific preferences or requirements but also limits the ease of subsequent manipulation of the resulting objects.

Motivated by the recent achievements in text-to-image generative models~\cite{rombach2022high, saharia2022photorealistic, kang2023scaling, tao2023galip}, several studies~\cite{wang2022clip, liu2022towards, wei2023taps3d} seek to emulate this success by conditioning on particular textual prompts. As a pioneer in the domain of text-guided 3D generation, Text2Shape~\cite{chen2019text2shape} introduces the first large-scale dataset of natural language descriptions for 3D furniture objects and combines conditional WGAN~\cite{arjovsky2017wasserstein} with 3D CNN to achieve supervised training for text-guided 3D object generation. This 3D dataset with human-annotated captions encourages various language-guided 3D generation methods~\cite{fu2022shapecrafter, mittal2022autosdf, liu2022towards, li2023diffusion}, especially with implicit 3D representations. These approaches directly learn the mapping between captions and corresponding 3D shapes, enabling the production of objects that align closely with specific details mentioned in the input text. While the inclusion of human-defined text supervision enhances the correlation between text prompt and generated shape, the scarcity of human-captioned 3D datasets for various object types (beyond furniture) restricts their applicability to specific object classes. Consequently, accurately aligning text descriptions with resulting 3D objects remains a challenging task.

To mitigate reliance on human-annotated datasets and achieve unsupervised text-to-3D generation, various methods~\cite{lin2023magic3d, metzer2023latent, poole2022dreamfusion, sanghi2022clip, sanghi2023clip, wei2023taps3d, huang2023textfield3d} leverage pre-trained text-driven 2D image synthesis network~\cite{saharia2022photorealistic, rombach2022high} or large vision and language models~\cite{li2022blip, li2023blip, dai2023instructblip, radford2021learning} to address the inherent modality difference between text and vision element. For instance, some approaches~\cite{lin2023magic3d, poole2022dreamfusion, metzer2023latent} draw guidance from powerful 2D diffusion models~\cite{saharia2022photorealistic, rombach2022high} by aligning the rendered RGB images from a random initialized NeRF~\cite{mildenhall2021nerf} with text-guided 2D diffusion priors, ensuring the optimized NeRF corresponds accurately to the textual description. Despite enabling zero-shot generation, additional optimization processes of these methods significantly increase the inference time as noted in~\cite{lin2023magic3d, poole2022dreamfusion}, which limits real-time responsiveness to user inputs.

Conversely, visual-language-based (V-L-based) methods~\cite{sanghi2022clip, sanghi2023clip, wei2023taps3d, huang2023textfield3d} tackle this challenge by training a latent generator using either rendered image embeddings or pseudo text embedding encoded by CLIP~\cite{radford2021learning}. Leveraging the aligned latent space of vision and text of CLIP, these methods can generate the correct latent for shape generation based on text prompt embeddings. While current V-L-based methods ease the necessity of paired 3D shapes with captions, they primarily generate shapes and textures at a general class and color level due to utilizing global (sentence) features of input texts as guidance for generating 3D objects. This implies the potential loss of detailed information in the text prompts, leading to similar shapes and textures generated from different fine-grained descriptions.

In this paper, we propose TriPlane Attention 3D Generator (TPA3D), a GAN-based text-guided 3D object generation network. Inspired by the unconditional GAN-based model of GET3D~\cite{gao2022get3d}, our TPA3D only utilizes 3D objects and their rendered images for generating high-fidelity 3D textured mesh. With text features extracted from the pre-trained CLIP text encoder, our proposed TriPlane Attention (TPA) block performs sentence and word feature refinement for the geometry and texture triplanes, allowing fine-grained 3D textured mesh to be produced via generator-based modules. We note that our TPA3D trains generators and discriminators in an unsupervised setting, performing instantly generation of high-fidelity 3D textured triplane corresponding to the detailed description without the human-annotated text-3D pairs for training supervision. This also makes GAN-based generation methods~\cite{huang2023textfield3d, wei2023taps3d} preferable over diffusion-based models~\cite{lin2023magic3d, poole2022dreamfusion} which require significant optimization costs.

We now highlight the contributions of this work below:

\begin{itemize}
    \item We propose a GAN-based network, named TriPlane Attention 3D Generator (TPA3D), performing sentence and word-level refinements of triplane features for fast text-guided 3D textured mesh generation.
    \item Our TriPlane Attention (TPA) performs plane-wise self-attention, cross-plane attention, and cross-word attention, allowing us to preserve intra-plane consistency, enhance 3D spatial connectivity, and integrate fine-grained information from the input text prompt for producing triplane features.
    %preserve intra-plane consistency and 3D spatial connectivity for producing triplane features.
    \item We demonstrate our method outperforms the state-of-the-art GAN-based text-to-3D method in various evaluation metrics, and has better textual alignment than SDS-based methods.
\end{itemize}

%% file: sec/2_related_works.tex
\section{Related Works}
\label{sec:related_works}

\subsection{Text-Guided 2D Image Synthesis}
With the advent of large-scale datasets~\cite{schuhmann2022laion, jia2021scaling} containing text-image pairs, 2D text-guided generative models are developed, which leverage direct supervision through RGB images and their corresponding captions. To retrieve desirable information from the text prompt, recent methodologies~\cite{ding2021cogview, ramesh2021zero, saharia2022photorealistic, rombach2022high} adopt attention mechanisms~\cite{vaswani2017attention} as a key strategy to integrate fine-grained text features into their designs. Specifically, auto-regressive-based models~\cite{ding2021cogview, ramesh2021zero} utilize transformers~\cite{chen2020generative} to establish connections between text tokens and image patch tokens. Likewise, Diffusion Models such as Imagen~\cite{saharia2022photorealistic} incorporate multiple cross-attention layers within the encoder and decoder, facilitating the learning of the denoising process. To enhance the resolution of the generated images, the Latent Diffusion Model (LDM)~\cite{rombach2022high} suggests shifting the denoising process to the latent space rather than the pixel level. 

However, the substantial computation cost incurred during inference still poses the interactivity concern~\cite{kang2023scaling}. GigaGAN~\cite{kang2023scaling} thus delves into the prospect of upscaling conditional StyleGAN~\cite{karras2020analyzing} to accommodate large-scale datasets~\cite{laine2020modular}. %The aim is to uphold the efficacy of generation and latent editing functionalities such as interpolation and style mixing. 
It further integrates cross-attention layers within the generator, focusing on both textual and visual features. While enabling the extraction of local details from comprehensive captions, the extension of these 2D methodologies to a 3D context remains an intricate challenge.

\subsection{Text-Guided 3D Object Generation}
Based on the success of text-guided image synthesis, numerous works~\cite{chen2019text2shape, liu2022towards, sanghi2022clip, sanghi2023clip, mittal2022autosdf, li2023diffusion, poole2022dreamfusion, lin2023magic3d, wei2023taps3d} thrive in developing approaches for text-guided 3D object generation. Notably, Text2Shape~\cite{chen2019text2shape} pioneers the text-to-3D domain by introducing the first large-scale 3D dataset with human-annotated captions for 3D furniture objects in ShapeNet~\cite{chang2015shapenet} and optimizing a conditional WGAN~\cite{arjovsky2017wasserstein} through supervised training. Subsequent studies~\cite{liu2022towards, mittal2022autosdf, li2023diffusion} adopt similar supervisory strategies to design text-to-3D networks. For instance, TITG3SG~\cite{liu2022towards} uses Implicit Maximum Likelihood Estimation (IMLE)~\cite{li2018implicit} as the latent generator, which minimizes the similarity between the ground truth latent vector and the most similar generated latent vector from a set of generated results. AutoSDF~\cite{mittal2022autosdf} utilizes a VQ-VAE~\cite{van2017neural} to encode 3D objects and updates an additional auto-regressive Transformer~\cite{vaswani2017attention} with text-3D pairs in the latent space during training to achieve text-guided latent vector generation. While these methods successfully achieve text-driven 3D object generation, the scarcity of human-annotated 3D datasets confines the applicability of these methods to specific classes. Consequently, recent endeavors share a common objective of reducing dependence on 3D datasets with human-define captions.

Recent advancements in cross-modality models~\cite{li2022blip, li2023blip, dai2023instructblip, radford2021learning} have been observed. Pre-trained on large-scale image and text-paired data, these models have become prominent tools for bridging natural language with visual elements. Consequently, several studies~\cite{poole2022dreamfusion, lin2023magic3d, wei2023taps3d, sanghi2022clip, sanghi2023clip} are exploring the potential of leveraging knowledge from these large language and vision models for text-guided 3D generation. Among these approaches, Magic3D~\cite{lin2023magic3d} and DreamFusion~\cite{poole2022dreamfusion} utilize diffusion models~\cite{rombach2022high} to align rendered RGB images from a random initialized NeRF with text-guided 2D diffusion priors. By capturing fine-grained information from word-level features, these methods can ensure the generated shape corresponds to the specific requirements in the text prompt. Despite successfully achieving zero-shot text-guided 3D generation, additional optimization is typically required as reported in~\cite{lin2023magic3d, poole2022dreamfusion}. Such extra computation efforts would limit their practicality as 3D generation tools.

To address the above issue, alternative approaches~\cite{wei2023taps3d, wang2022clip, sanghi2022clip, sanghi2023clip} aim to capitalize on the aligned vision and language latent space of CLIP~\cite{radford2021learning} to generate text-conditioned latent for 3D objects generation. For instance, CLIP-Forge~\cite{sanghi2022clip} introduces a flow-based model to learn the mapping between CLIP image embedding of rendered RGB image and the 3D shape latent. Leveraging the aligned text-image latent space, the latent generator predicts the suitable shape latent based on the text features of user input. Conversely, TAPS3D~\cite{wei2023taps3d} introduces a novel captioning module to identify the best pseudo caption for the rendered images of 3D objects in the training set by maximizing the CLIP score between each image and the formulated text. By obtaining suitable pseudo captions for the 3D objects, TAPS3D directly fine-tunes a pre-trained conditional 3D generator~\cite{gao2022get3d} with the supervision of generated pseudo captions. Although the above technique does not require human-defined captions, their reliance on CLIP global text embedding as the primary guidance constrains their ability to precisely generate 3D objects matching detailed text inputs.

%% file: sec/3_preliminary.tex
\section{Preliminary}
\label{sec:preliminary}
\input{figure/model_overview}
For the sake of clarity, we provide a brief review of GET3D~\cite{gao2022get3d}, which generates textured 3D shapes and serves as the generator backbone for our method. The model of GET3D is a GAN-based single-class unconditional 3D generator. Adapted from the generator of StyleGAN2~\cite{karras2020analyzing}, GET3D first maps random noises $\mathbf{z}_{\text{geo}}\in \mathcal{N}(0, I)$ and $\mathbf{z}_{\text{tex}}\in \mathcal{N}(0, I)$ to latent vectors $\mathbf{w}_{\text{geo}}\in\mathbb{R}^{512}$  $\mathbf{w}_{\text{tex}}\in\mathbb{R}^{512}$. Subsequently, for the $i$-th layer of the generator, $\mathbf{w}_{\text{geo}}$ and $\mathbf{w}_{\text{tex}}$ are utilized to control the generation of the geometry triplane $c^{g}_i\in\mathbb{R}^{H_i\times W_i\times (3d) }$ and the texture triplane $c^{t}_i\in\mathbb{R}^{H_i\times W_i\times (3d) }$, where $d$ is the feature dimension, $H_i$ and $W_i$ denote the size of the triplane at that layer. After summing up all $c^g_i$ and $c^t_i$, the final triplanes $f_{N}^{g}=\sum_{i=1}^N(c^g_i)$ and $f_{N}^{t}=\sum_{i=1}^N(c^t_i)$ are fed to DMTet~\cite{shen2021deep} to generate the output 3D textured mesh. In order to sum up all triplanes, a bilinear upsampling strategy is applied here (except for the triplanes with a size equal to $H_N \times W_N$) to align the resolution of all triplanes. Please refer to the supplementary materials for the details.

To train the generator of GET3D via 2D supervision, a differentiable renderer~\cite{laine2020modular} is utilized to render the generated 3D textured mesh into 2D RGB image $I_{\text{fake}}$ and silhouette mask $M_{\text{fake}}$. Following the discriminator architecture of StyleGAN~\cite{karras2019style}, GET3D applies two separate discriminators conditioned on the camera pose for the RGB and the mask images, respectively. 

With the advantage of instant generation of high-fidelity textured shapes, our method is built on top of GET3D with the modification for text-guided 3D object generation. Specifically, to generate a textured mesh that aligns with the detailed description from the input text, our approach performs text-guided refinement on triplanes $c^g_i$ and $c^t_i$ by applying word-level information. With this modification, our model is able to generate high-fidelity 3D textured meshes while ensuring precise correspondences to the fine-grained textual conditions.

%% file: figure/model_overview.tex
\begin{figure}[tp]
  \centering
  \includegraphics[width=0.92\linewidth]{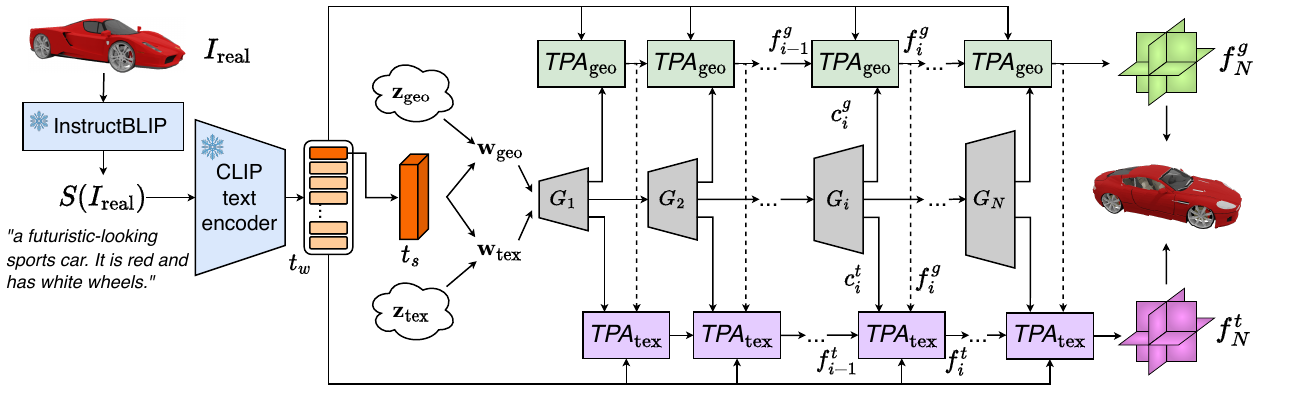}
  %\vspace{-1mm}
  \caption{\textbf{Overview of TPA3D for fast text-guided 3D generation.} By taking sentence and word-level features $t_s$ and $t_w$ as the inputs, TPA3D utilizes generator $G$ and triplane attention (TPA) modules to predict the associated triplane features for 3D textured mesh generation, with 3D content information properly observed. Following GET3D~\cite{gao2022get3d}, each $G$ contains branches for geometry and texture synthesis. Note that InstructBLIP~\cite{dai2023instructblip} is applied to produce pseudo captions from rendered images during training, while CLIP~\cite{radford2021learning} extracts the resulting text features. And, $c_i$ and $f_i$ denote the sentence and word-level triplane features at each layer $i$, respectively.}
  %\vspace{-1mm}
  \label{fig:model_overview}
\end{figure}

%% file: sec/4_methods.tex
\section{Method}
\label{sec:methods}

\subsection{Problem Formulation and Model Overview}

We first define the problem definition and the notation used in this paper. Given an input text prompt $S$ describing desirable information of an object, our goal is to generate a 3D textured mesh that matches $S$ \textit{without} reliance on human-defined text-3D pairs for training supervision. To achieve this goal, we propose a novel deep learning framework of TriPlane Attention 3D Generator (TPA3D). 

As depicted in Figure~\ref{fig:model_overview}, our TPA3D contains two network modules. First, we have \textit{sentence-level triplane generators} $G=\{G_1,...,G_N\}$ for generating sentence-level triplanes $c^g_i$ and $c^t_i$, with latent vectors $\mathbf{w}_{\text{geo}}$ and $\mathbf{w}_{\text{tex}}$ derived from sentence features $t_s$. The other module is \textit{TriPlane Attention block} (TPA$_{\text{geo}}$ and TPA$_{\text{tex}}$), refining $c^g_i$ and $c^t_i$ into word-level triplanes $f^g_i$ and $f^t_i$ with word features $t_w$. Given an input text $S$, we apply a pre-trained CLIP text encoder~\cite{radford2021learning} to convert $S$ into a sentence feature $t_{s}\in\mathbb{R}^{512}$ and word features $t_{w}\in\mathbb{R}^{77\times 512}$ as global and detailed text information, respectively. We concatenate $t_s$ to randomly sampled noises $\mathbf{z}_{\text{geo}}$ and $\mathbf{z}_{\text{tex}}$ as conditions to generate latent vectors $\mathbf{w}_{\text{geo}}$ and $\mathbf{w}_{\text{tex}}$ for $G$ to produce sentence-level geometry triplane $c^{g}_i$ and texture triplane $c^{t}_i$ respectively from each layer $G_i$. To further enhance the consistency and connectivity of the generated triplanes and incorporate word-level information to capture geometric and textured details, the proposed novel TPA blocks are adopted to refine $c^g_i$ and $c^t_i$ and derive word-level geometry triplane $f^g_i$ and texture triplane $f^t_i$. Finally, we follow GET3D~\cite{gao2022get3d} to produce the 3D textured mesh from $f^g_N$ and $f^t_N$ via DMTet~\cite{shen2021deep} as noted in Sect.~\ref{sec:preliminary}. 
We now provide a detailed explanation of our TPA3D in the following subsections.

\subsection{Pseudo Caption Generation}
As discussed in Sect.~\ref{sec:related_works}, traditional text-guided 3D generation approaches~\cite{liu2022towards, mittal2022autosdf, chen2019text2shape} require human-annotated text-3D pairs to enable supervised training. To mitigate the reliance on human-annotated text-3D pairs, we leverage a pre-trained image captioning model of InstructBLIP~\cite{dai2023instructblip} to produce the detailed pseudo caption $S(I_{\text{real}})$ for a rendered image $I_{\text{real}}$ of its 3D version, providing pseudo text-3D pairs for training. As suggested by~\cite{conceptual_captions}, it is necessary to remove redundant phrases from generated captions to mitigate potential distraction for 3D generation (e.g., \textit{``in the image''}, \textit{``This is a 3D model''}, and \textit{``black background''}). Please refer to supplementary materials for details of this filtering step. With such pseudo caption $S(I_{\text{real}})$ and rendered image $I_{\text{real}}$ pairs obtained, our TPA3D is subsequently designed to accommodate and leverage this detailed description.

\subsection{Triplane Attention 3D Generator}

We now detail how TPA3D realizes text-guided 3D generation, with the ability to produce 3D content with desirable shape and texture information. Given the input pseudo caption $S(I_{\text{real}})$, we apply the CLIP text encoder~\cite{radford2021learning} to extract sentence features $t_s$ and word features $t_w$. For $t_s$, we directly use the output CLIP text embeddings~\cite{ramesh2022hierarchical}. As for $t_w$, we follow~\cite{saharia2022photorealistic, kang2023scaling} and extract the features from the second-last layer of the CLIP text encoder.

\subsubsection{Sentence-Level Triplane Generator.}
As illustrated in Figure~\ref{fig:model_overview}, the sentence-level triplane generator $G$ generates sentence-level triplanes $c^g_i\in\mathbb{R}^{H_i\times W_i\times (3d)}$ and $c^t_i\in\mathbb{R}^{H_i\times W_i\times (3d)}$ with latent vectors $\mathbf{w}_{\text{geo}}$ and $\mathbf{w}_{\text{tex}}$, which are conditioned on sentence features $t_s$, at each layer $G_i$. Following the generator architecture of GET3D~\cite{gao2022get3d}, the sentence-level generator $G$ takes geometry latent vector $\mathbf{w}_{\text{geo}}$ and texture latent vector $\mathbf{w}_{\text{tex}}$ as inputs to generate the geometry triplane $c^g_i$ and the texture triplane $c^t_i$ at each layer $G_i$. For each layer, $G_i$ takes the layer features of $G_{i-1} $ as the input, and uses modulated convolution layers~\cite{karras2020analyzing} with $\mathbf{w}_{\text{geo}}$ as style information to generate the layer features, which will be propagated to the next layer $G_{i+1}$ for the subsequent generation. To generate triplanes at each layer $G_i$, two additional modulated convolution layers are applied to layer features and take $\mathbf{w}_{\text{geo}}$ and $\mathbf{w}_{\text{tex}}$ as styles to generate the sentence-level geometry triplane $c^g_i$ and the texture triplane $c^t_i$ respectively. Because $\mathbf{w}_{\text{geo}}$ and $\mathbf{w}_{\text{tex}}$ are conditioned on the sentence features $t_s$, $c^g_i$ and $c^t_i$ only contain sentence-level information for textured mesh generation. 

\input{figure/TPA}

\subsubsection{Word-Level Triplane Refinement via TPA.}
\label{sec:TPA}

To further refine the above sentence-level triplanes with detailed information matching the text input, we propose the TriPlane Attention (TPA) block that performs word-level refinement to generate word-level geometry triplane $f^g_i\in\mathbb{R}^{H_i\times W_i\times (3d)}$ and texture triplane $f^t_i\in\mathbb{R}^{H_i\times W_i\times (3d)}$ accordingly. As depicted in Figure~\ref{fig:TPA}, for a TPA$_{\text{geo}}$ at layer $i$, we take the sentence-level geometry triplane $c_i^g$ generated by $G_i$ and the output $f_{i-1}^g$ of the TPA$_{\text{geo}}$ in the previous layer to obtain three plane input features $f_{i,xy}^g, f_{i,yz}^g, f_{i,xz}^g \in\mathbb{R}^{H_i\times W_i\times d}$. For better understanding, we use $\langle \cdot \rangle$ as an operator to stack the feature channel (i.e., $\langle f^g_{i,xy}\cdot f^g_{i,yz}\cdot f^g_{i,xz}\rangle \in\mathbb{R}^{H_i\times W_i\times (3d)}$). Therefore, $\langle f^g_{i,xy}\cdot f^g_{i,yz}\cdot f^g_{i,xz}\rangle $, the input of the TPA$_{\text{geo}}$ is defined as,
\begin{equation}
  \langle f^g_{i,xy}\cdot f^g_{i,yz}\cdot f^g_{i,xz}\rangle = f_{i-1}^{g} + c_i^g.
\end{equation}
As for TPA$_{\text{tex}}$, we additionally include the output $f^g_i$ from TPA$_{\text{geo}}$ with a weight $\alpha=0.5$ as the input of TPA$_{\text{tex}}$ to generate texture matches the corresponding geometry. Therefore, the input of TPA$_{\text{tex}}$ is defined as:
\begin{equation}
  \langle f^t_{i,xy}\cdot f^t_{i,yz}\cdot f^t_{i,xz}\rangle = f_{i-1}^{t} + c_i^t + \alpha * f_i^g.
\end{equation}
Note that the model architectures of TPA$_{\text{geo}}$ and TPA$_{\text{tex}}$ are the same except for their inputs.\\

\textbf{Triplane-Feature Consistency and Connectivity.}
To ensure proper modeling of sentence-level object information through triplane features before incorporating word-level fine-grained details, our TPA (taking TPA$_{\text{geo}}$ for example) in Figure~\ref{fig:TPA} initiates by prioritizing the acquisition of intra-plane consistency. Subsequently, additional efforts can be placed on fostering inter-plane 3D spatial connectivity across all planes, establishing a foundation for subsequent refinement processes.
To inject sentence-level content pertaining to the desired object, we start from representation learning for each triplane. As depicted in the lower middle of Figure~\ref{fig:TPA}, this is achieved by observing intra-plane consistency for each triplane feature. That is, we choose to perform plane-wise self-attention on each plane feature to extract plane-wise content features. To further enhance the 3D spatial information inherent in our plane-wise content features and ensure comprehensive multi-aspect correspondence across different planes, we then employ a cross-plane attention mechanism to establish inter-plane connectivity. This involves treating the fused triplane feature $f^g_{i,p}$ (i.e., the output of applying self-attention on the concatenation of triplane features $f^g_{i,xy}, f^g_{i,yz},$ and $f^g_{i,xz}$) as both key and value, while employing the three plane-wise content features as queries to execute attention operations.
% self-attention on the concatenation 
%  of triplane features $(f^g_{i,xy}, f^g_{i,yz}, f^g_{i,xz})$ to enhance the inter-plane connectivity} between each plane and concatenate the output triplane as \ben{spatial features $f^g_{i,p}$}. 
% % \ben{To combine the intra-plane consistency and the inter-plane connectivity,}
% we fuse three plane-wise content features with $f^g_{i,p}$ separately and perform cross-plane attention. More precisely, we use $f^g_{i,p}$ as key and value, three plane-wise content features as the queries for performing attention operation. 
We note that to decrease the number of parameters, the weights of cross-plane attention and plane-wise self-attention are shared between each plane feature. Finally, we concatenate three output plane features as self-refined features $f^g_{i,q}$ for later fine-grained word-level refinement.\\

\textbf{Refinement with Word Features.}
To incorporate word-level information into triplane features for 3D generation, we perform the word-level refinement by cross-word attention. As shown in the right of TPA in Figure~\ref{fig:TPA}, the cross-word attention takes self-refined features $f^g_{i,q}$ as query, word features $t_w$ as key and value, to refine $f^g_{i,q}$ to the output word-level triplane $f_i^g$. Therefore, $f_i^g$ will include 3D spatial information and word-level information. With the refinement by TPA$_{\text{geo}}$ blocks and TPA$_{\text{tex}}$ blocks, we obtain final word-level triplanes $f^g_N$ and $f^t_N$, which are used to generate the 3D textured mesh that matches the detailed description. Different from DiffTF~\cite{cao2023large}, our TPA blocks have additional cross-word attention to utilize word features, so the generated triplanes contain fine-grained information corresponding to the detailed text prompt.

\subsection{Text-Guided Discriminators}

To train our TPA3D, we deploy and train the discriminators conditioned on the text inputs. Following GET3D~\cite{gao2022get3d}, we use the same architectures of two discriminators $D_{\text{rgb}}$ and $D_{\text{mask}}$ for RGB images and masks, respectively. To properly design text-guided discriminators, we concatenate the sentence features $t_s$ to the camera pose condition as a new condition. In this case, the discriminators not only need to know whether the input rendered image is real or fake, but also have to judge whether the image matches the given detailed caption. For $D_{\text{rgb}}$ and $D_{\text{mask}}$, the adversarial objective is formulated as,
\begin{equation}
\begin{split}
  \mathcal{L}(D_{\text{rgb}}, G) & = \mathbb{E}_{{t_s}\in T}\ g(D_{\text{rgb}}(I_{\text{fake}}, t_s)) \\ 
  & + \mathbb{E}_{{t_s}\in T,I_{\text{real}}\in p_{\text{rgb}}}\ (g(-D_{\text{rgb}}(I_{\text{real}}, t_s)) \\
  & + \lambda \lVert \nabla D_{\text{rgb}}(I_{\text{real}}) \rVert^2_2),
\end{split}
\end{equation}
\vspace{-1mm}
\begin{equation}
\begin{split}
  \mathcal{L}(D_{\text{mask}}, G) & = \mathbb{E}_{{t_s}\in T}\ g(D_{\text{mask}}(M_{\text{fake}}, t_s)) \\ 
  & + \mathbb{E}_{{t_s}\in T,M_{\text{real}}\in p_{\text{mask}}}\ (g(-D_{\text{mask}}(M_{\text{real}}, t_s)) \\
  & + \lambda \lVert \nabla D_{\text{mask}}(M_{\text{real}}) \rVert^2_2),
\end{split}
\end{equation}
where $g(x)=-\log(1+\exp(-x))$. Note that $p_{\text{rgb}}$ and $p_{\text{mask}}$ represent the distributions of real rendered RGB images and silhouette masks, and $\lambda$ is a hyperparameter.

To introduce additional discriminative ability during training, we use additional negative pairs in the mismatching objective $\mathcal{L}_{\text{mis}}$ to make the model more sensitive to mismatched text conditions. Therefore, the mismatching objective is formulated as, 
\begin{equation}
\begin{split}
  \mathcal{L}_{\text{mis}} & = \mathbb{E}_{{t'_s}\in T'}\ g(D_{\text{rgb}}(I_{\text{fake}}, t'_s)) \\
                    & + \mathbb{E}_{{t'_s}\in T', I_{\text{real}}\in p_{\text{rgb}}}\ g(D_{\text{rgb}}(I_{\text{real}}, t'_s)) \\
                    & + \mathbb{E}_{{t'_s}\in T'}\ g(D_{\text{mask}}(M_{\text{fake}}, t'_s)) \\
                    & + \mathbb{E}_{{t'_s}\in T', M_{\text{real}}\in p_{\text{mask}}}\ g(D_{\text{mask}}(M_{\text{real}}, t'_s)),
\end{split}
\end{equation}
where $T'$ denotes the set of mismatched sentence features.

\subsection{Training and Inference}
\subsubsection{Training.}
Since we use InstructBLIP~\cite{dai2023instructblip} to generate detailed pseudo captions, we only require rendered image $I_{\text{real}}$ of the 3D object as our training data. As a result, the pseudo caption $S(I_{\text{real}})$, sentence features $t_s$, and word features $t_w$ can be produced as described above. The generator is trained to generate $I_{\text{fake}}$ and feed $I_{\text{real}}$ and $I_{\text{fake}}$ into the discriminators. To stabilize the training process, we use an additional CLIP similarity score for $I_{\text{fake}}$ and $t_s$ as a training objective $\mathcal{L}_{\text{clip}}$. Therefore, the overall training objective is defined as:
\begin{equation}
  \mathcal{L} = \mathcal{L}(D_{\text{rgb}}, G) + \mathcal{L}(D_{\text{mask}}, G) + \mathcal{L}_{\text{mis}} + \mathcal{L}_{\text{clip}}.
\end{equation}
\subsubsection{Inference.}
For inference, one can replace the generated pseudo caption directly with the desirable input text prompt, fed into the generator for synthesizing the fine-grained 3D textured mesh that matches the input text prompt.

%% file: figure/TPA.tex
\begin{figure}[tp]
    \centering
    \includegraphics[width=0.8\linewidth]{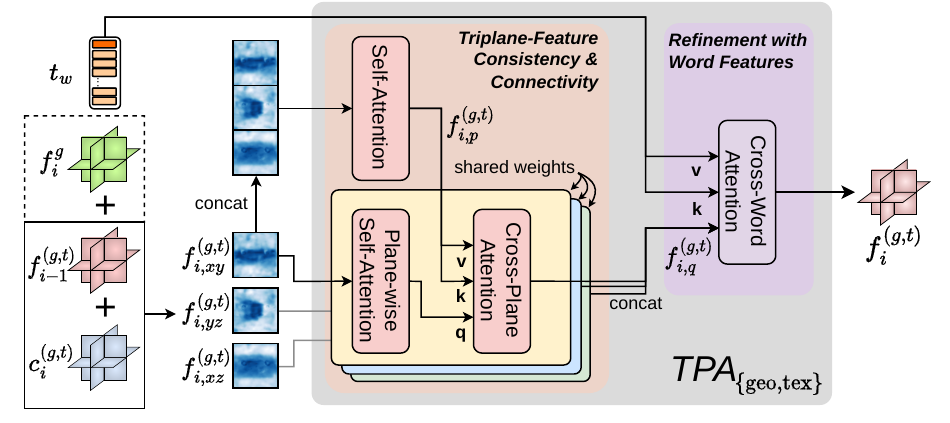}
    %\vspace{-1mm}
    \caption{\textbf{Design of TriPlane Attention (TPA).} TPA first performs plane-wise self-attention and cross-plane attention to 3D triplane features to enforce intra-plane consistency and 3D spatial connectivity, respectively. Cross-word attention is subsequently performed to exploit word-level features for incorporating detailed information. Note that for TPA$_{\text{tex}}$, additional geometry triplane features $f^g_i$ are included to incorporate geometry information for texture generation.}
    %\vspace{-1mm}
    \label{fig:TPA}
\end{figure}

%% file: sec/5_experiments.tex
\section{Experiments}
\label{sec:experiments}

\input{table/fid_clip_r}
\subsection{Dataset}

We train and evaluate our models on the synthetic 3D ShapeNet~\cite{chang2015shapenet} dataset following ~\cite{gao2022get3d, wei2023taps3d, cheng2023sdfusion, xu2023dream3d} and further include real-scanned 3D dataset OmniObject3D~\cite{chang2015shapenet} to demonstrate its applicability on versatile real-world data. Specifically, we choose categories with diverse geometric and textural details, including \textit{Car, Chair}, and \textit{Motorbike} for ShapeNet. Since the number of objects for a single class in OmniObject3D is much smaller than ShapeNet, we combine \textit{Toy Bus, Toy Car, Toy Truck,} and \textit{Toy Train} as \textit{Vehicle}, and combine \textit{Gloves, Hat, Helmet,} and \textit{Shoes} as \textit{Accessory} for OmniObject3D.
In our experiments, we generate images with a high resolution of 1024$\times$1024 by rendering each textured shape from 24 randomly sampled camera angles. For categories with fewer shapes, like \textit{Motorbike, Vehicle,} and \textit{Accessory}, we increase the view count to 100 to ensure a comparable volume of rendered images.
Finally, for fair comparison purposes, we generate \textit{one} pseudo caption for each rendered image for quantitative and qualitative evaluation.

\input{figure/compare}

\subsection{Quantitative Results}

To quantitatively evaluate the capability of our method, we compare our proposed TPA3D with several existing state-of-the-art works, including GET3D~\cite{gao2022get3d} and TAPS3D~\cite{wei2023taps3d}, with the following evaluation metrics. To evaluate the fidelity and quality of generated shapes, we render 3D textured shapes into RGB images from 24 random camera views and compute Fréchet inception distances (FID)~\cite{heusel2017gans} of rendered images. As for the evaluation of consistency between text prompts and generated objects, we adopt the CLIP R-precision@5~\cite{park2021benchmark} as our main metric. Lastly, all the models are trained and evaluated with pseudo captions generated by InstuctBLIP~\cite{dai2023instructblip}. The results of FID, as presented in Table~\ref{table:fid}, reveal that our TPA3D achieves comparable scores in FID with state-of-the-art 3D generator GET3D (as the performance upper bound for ShapeNet~\cite{chang2015shapenet}) and outperforms text-guided 3D generation method TAPS3D in all classes. 
We note that, while GET3D excels in producing high-quality shapes, it is limited to unconditional shape generation and is hard to deal with high-diversity datasets composed of multiple classes without further guidance, and thus our TPA3D achieves higher scores in FID than GET3D on OmniObject3D~\cite{wu2023omniobject3d}.
%Notably, while GET3D excels in producing high-quality shapes, it is limited to unconditional shape generation and \ben{lacks the ability to deal with high-diversity datasets (i.e., \textit{Vehicle}).} 
As for the correspondence between input texts and generated objects, Table~\ref{table:clip_r_precision} demonstrates our TPA3D achieves higher CLIP R-precision across all classes compared to TAPS3D, which only utilizes sentence features. The result indicates that our generated shape better aligns with specified input text conditions. This validates the effectiveness of our approach, which leverages word features to enhance details in triplane features, resulting in our generated shape and texture retrieving detailed requirements specified in the text prompt.

\input{figure/qualitative}
\subsection{Qualitative Results}

 To qualitatively evaluate the ability to deal with detailed descriptions, we first compare our TPA3D with TAPS3D~\cite{wei2023taps3d}. The results in Figure~\ref{fig:compare_taps3d} demonstrate that our model produces shapes accurately aligned to the text prompt while TAPS3D only comprehends simple modifiers and affects the output shape with different colors. For ShapeNet~\cite{chang2015shapenet} (see the third column of Figure~\ref{fig:compare_shapenet}), TAPS3D only generates a wooden chair and ignores further details provided in the text prompt. In contrast, our TPA3D accurately captures all details such as \textit{``rounded back''}, \textit{``armrests''}, and \textit{``linen seat''}. For OmniObject3D~\cite{wu2023omniobject3d} (see the second column of Figure~\ref{fig:compare_omni}), TAPS3D mixes the colors and misunderstands the accurate sub-class. In contrast, our TPA3D separates the colors \textit{``green''} and \textit{``white''}, and constructs the shape of \textit{``garbage truck''}. This qualitatively verifies the effectiveness of our design of incorporating word-level triplane refinement in our TPA blocks. Furthermore, we present additional qualitative results for text-guided 3D generation in Figure~\ref{fig:brand_color} with different combinations of color and subclass. In this figure, we observe that our TPA3D exhibits impressive precision in generating textured shapes aligned with various combinations of subclass and color provided in text prompts. By fixing the random seed and subclass in each column, we can also observe that TPA3D only modifies textures with the changed color and maintains nearly identical shapes. This further verifies the effectiveness of word-level refinement in TPA for disentangling geometry and texture information.

\subsection{Further Analysis}
\input{figure/manipulation}
\subsubsection{Text-Guided Manipulation.}
With the proper separation of geometry and texture triplane features, our TPA3D is able to manipulate the generated objects by simply changing the input text description and fixing the same random seed for sampling the initial noises $\mathbf{z}_{\text{geo}}$ and $\mathbf{z}_{\text{tex}}$. As shown in Figure~\ref{fig:manipulation}, we first generate a chair object via the input text \textit{``a wooden chair''}. By adding different text descriptions to the original one, our TPA3D manipulates the original chair accordingly without changing details unrelated to the additional descriptions. Such a manipulation property may improve its practicability as a 3D content creation tool for users to control the output incrementally.

\input{figure/sds_compare}
\subsubsection{Comparison with SDS-Based Methods.}
Since score distillation sampling (SDS) has shown significant performance in high-fidelity text-to-3D generation, we also compare TPA3D with SDS-based methods of DreamFusion~\cite{poole2022dreamfusion} and Magic3D~\cite{lin2023magic3d}. As shown in Figure~\ref{fig:sds_compare}, our method exhibits higher correspondence between complex input texts and generated objects. For example, in the fourth column of Figure~\ref{fig:sds_compare}, our TPA3D accurately separates the colors of \textit{``black office chair''} and \textit{``blue seat''}, while SDS-based methods either mix the colors or mismatch the colors to the parts of chairs. This is because, SDS-based methods heavily rely on pre-trained 2D diffusion models (e.g., Stable Diffusion~\cite{rombach2022high}), and thus they are not able to generate 3D objects directly from complex textual descriptions. Such limitations (i.e., dependence on 2D diffusion models) have been discussed in previous works such as StructureDiffusion~\cite{feng2022training} and Attend-and-excite~\cite{chefer2023attend}.

\subsubsection{Ablation Study on TPA.}
Since the proposed TPA module serves as the major technical component in TPA3D, we conduct several ablation studies to verify the design of TPA. In particular, we assess the design of TPA in three aspects: functions of cross-plane and cross-word attention in TPA, performances with different numbers of TPA blocks, and the improvement on the shape quality and the textual alignment with only TPA$_{\text{geo}}$ or TPA$_{\text{tex}}$. Due to page limitations, please refer to Sect.~\textcolor{red}{B}, Table~\textcolor{red}{A1}, and Table~\textcolor{red}{A2} in the supplementary materials for the complete ablation study results.

\input{table/inference_time}
\subsubsection{Inference Speed Comparison.}
To assess the real-time performance of each text-guided 3D generative model, we present the inference time of existing methods (reported from~\cite{wei2023taps3d, lin2023magic3d}) in Table~\ref{table:inference_time}. Notably, SDS-loss optimization-based approaches~\cite{lin2023magic3d, poole2022dreamfusion} require tens of minutes to complete the inference time optimization for each text input. In contrast, our proposed method maintains an instant inference speed comparable to GAN-based networks. Our approach achieves high-resolution image renderings at 1024$\times$1024 in just tens of milliseconds and generates textured meshes within three seconds, similar to the performance of other GAN-based generators such as TAPS3D~\cite{wei2023taps3d} and TITG3SG~\cite{liu2022towards}.
%This verifies our selection of a GAN-based generator backbone over diffusion-based ones.

%% file: table/fid_clip_r.tex
\begin{table}[tp]
    \centering
    \caption{\textbf{Quantitative results in terms of (a) FID$\downarrow$ and (b) CLIP R-Precision@5$\uparrow$.} Compared to TAPS3D with only sentence-level features, our TPA3D performs additional word-level refinement and results in better visual quality and improved alignment between generated shapes and given text prompts. Note that \textit{Acc.} represents \textit{Accessory} in tables.}
    \begin{subtable}[c]{0.495\textwidth}
        \centering
        \resizebox{1.05\textwidth}{!}{
            {
            \setlength{\tabcolsep}{3pt}
            \renewcommand{\arraystretch}{1.3}
            \begin{tabular}{l|ccc|cc}
            \toprule
            \multicolumn{1}{l|}{\multirow{2}{*}{Method}} & \multicolumn{3}{c|}{ShapeNet} & \multicolumn{2}{c}{OmniObject3D} \\
            \cline{2-6}
            \multicolumn{1}{l|}{} & Car & Chair & Motorbike & Vehicle & Acc. \\ \midrule \midrule
            GET3D~\cite{gao2022get3d} & \textbf{11.50} & \textbf{22.75} & \textbf{49.98} & \underline{98.15} & \underline{145.66} \\
            \midrule
            TAPS3D~\cite{wei2023taps3d} & 26.37 & 44.70 & 84.83 & 152.34 & 172.14 \\
            Ours (TPA3D) & \underline{18.50} & \underline{38.11} & \underline{77.69} & \textbf{68.80} & \textbf{83.31} \\
            \bottomrule
            \end{tabular}}
        }
        \caption{FID$\downarrow$}
        \label{table:fid}
    \end{subtable}
    \begin{subtable}[c]{0.495\textwidth}
        \centering
        \resizebox{1.0\textwidth}{!}{
        {
        \setlength{\tabcolsep}{3pt}
        \renewcommand{\arraystretch}{1.4}
        \begin{tabular}{l|ccc|cc}
            \toprule
            \multicolumn{1}{l|}{\multirow{2}{*}{Method}} & \multicolumn{3}{c|}{ShapeNet} & \multicolumn{2}{c}{OmniObject3D} \\
            \cline{2-6}
            \multicolumn{1}{l|}{} & Car & Chair & Motorbike & Vehicle & Acc. \\ \midrule
            TAPS3D~\cite{wei2023taps3d} & 12.55 & 7.52 & 5.00 & 9.47 & 6.67 \\
            Ours (TPA3D) & \textbf{80.94} & \textbf{38.58} & \textbf{24.76} & \textbf{65.26} & \textbf{64.44} \\
            \bottomrule
        \end{tabular}}}
        \caption{CLIP R-Precision@5$\uparrow$}
        \label{table:clip_r_precision}
    \end{subtable}
    %\vspace{-2mm}
\end{table}

%% file: figure/compare.tex
\begin{figure}[tp]
    \centering
    \begin{subfigure}{0.95\linewidth}
        \includegraphics[width=1\linewidth]{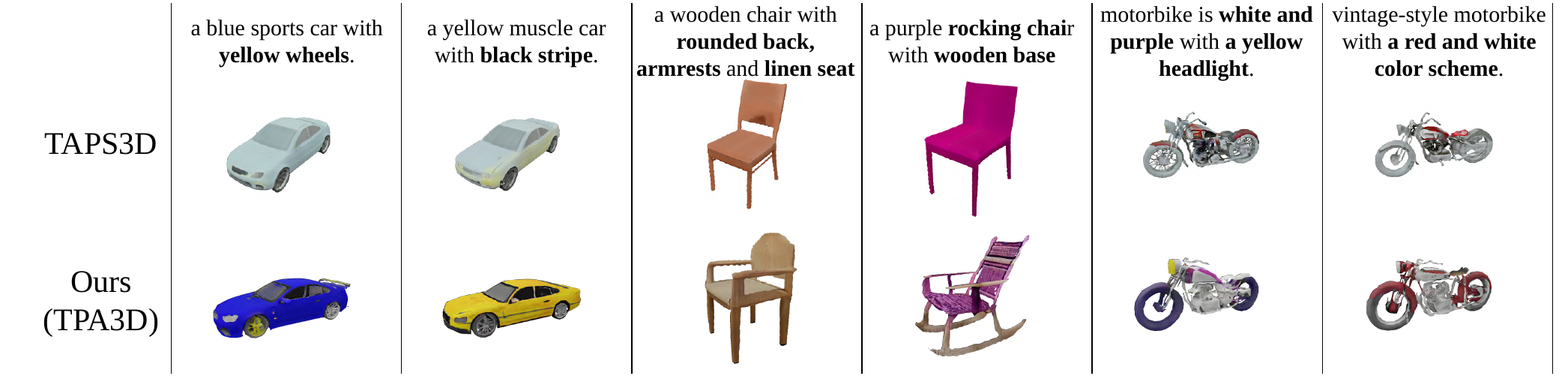}
        \caption{ShapeNet~\cite{chang2015shapenet}}
        \label{fig:compare_shapenet}
    \end{subfigure}
    \begin{subfigure}{0.95\linewidth}
        \includegraphics[width=1\linewidth]{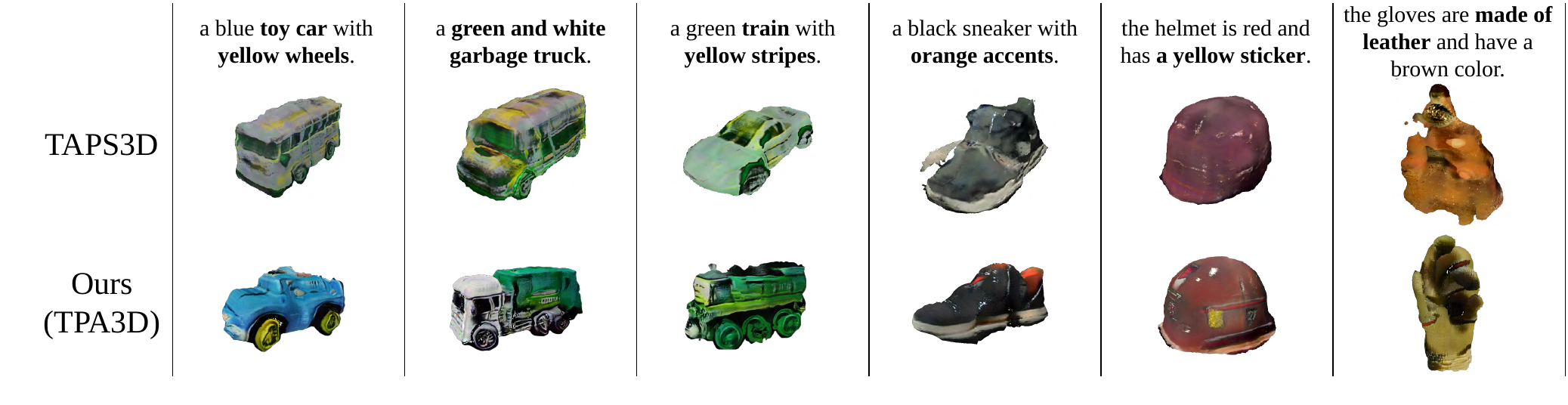}
        \caption{OmniObject3D~\cite{wu2023omniobject3d}}
        \label{fig:compare_omni}
    \end{subfigure}
    %\vspace{-5mm}
    \caption{\textbf{Qualitative comparisons with TAPS3D on (a) ShapeNet and (b) OmniObject3D.} Given detailed input prompts, our TPA3D generates accurate shapes aligned to prompts, while TAPS3D only realizes general classes and simple colors.}
    %\vspace{-6mm}
    \label{fig:compare_taps3d}
\end{figure}

%% file: figure/qualitative.tex
\begin{figure}[tp]
    \centering
    \includegraphics[width=0.85\linewidth]{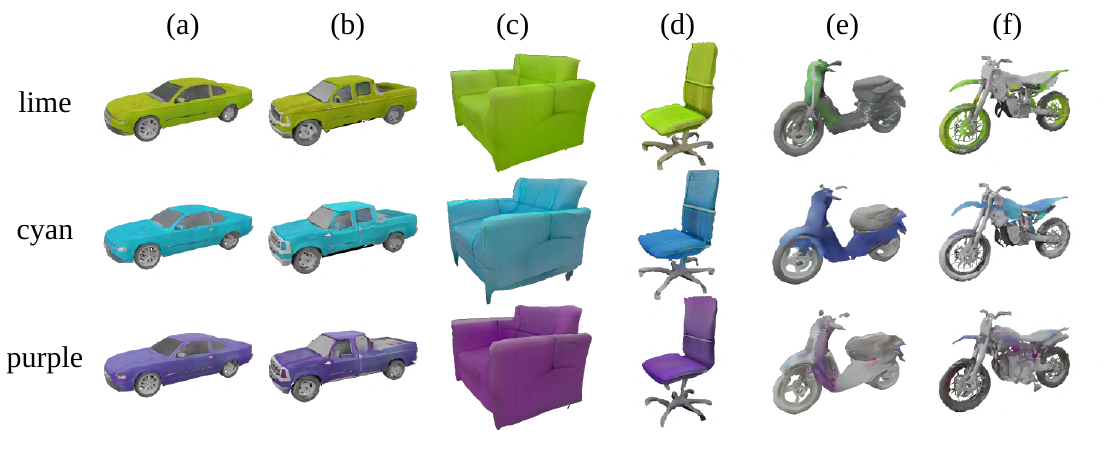}
    %\vspace{-3mm}
    \caption{\textbf{Example text-guided 3D generation results of TPA3D.} 
    We consider input prompts of ``a \{color\} \{object\}'' with multiple colors and sub-classes for generation. Each column stands for a different color, while each row stands for a unique sub-class: (a)\textit{``muscle car''} (b)\textit{``pickup truck''} (c)\textit{``sofa''} (d)\textit{``office chair''} (e)\textit{``scooter''} (f)\textit{``dirt bike''}. Note that the same seeds are applied for sampling $\mathbf{z}_{\text{geo}}$ and $\mathbf{z}_{\text{tex}}$ for each row.} 
    \label{fig:brand_color}
    % \vspace{-4mm}
\end{figure}

%% file: figure/manipulation.tex
\begin{figure}[!htb]
    \centering
    \includegraphics[width=0.95\linewidth]{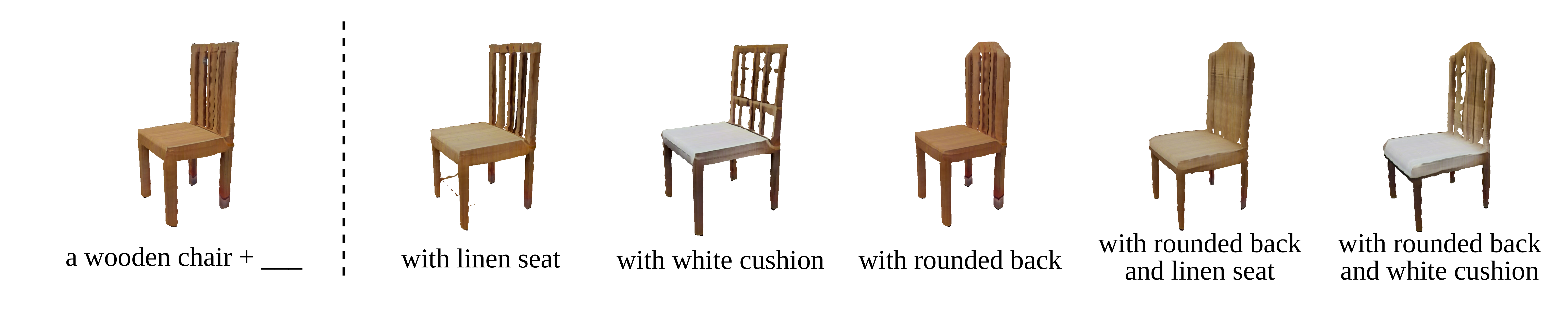}
    %\vspace{-7mm}
    \caption{\textbf{Examples of chair manipulation by adding different detailed text descriptions.} The left shows a chair generated from the input text \textit{``a wooden chair"}. With the same random seed for sampling $\mathbf{z}_{\text{geo}}$ and $\mathbf{z}_{\text{tex}}$, five distinct manipulations are produced by adding different detailed text descriptions.}
    \label{fig:manipulation}
    %\vspace{-5mm} % Adjust the spacing as needed
\end{figure}

%% file: figure/sds_compare.tex
\begin{figure}[tp]
    \centering
    \includegraphics[width=0.95\linewidth]{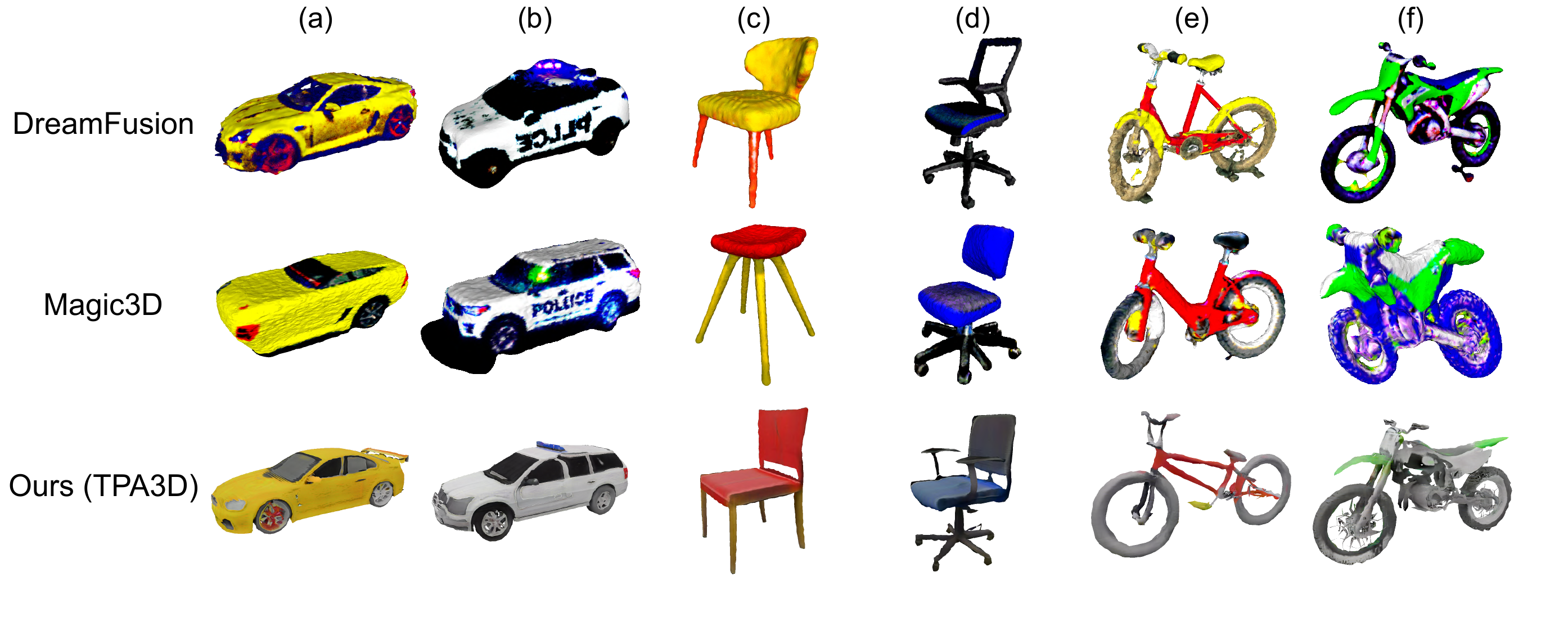}
    %\vspace{-8mm}
    \caption{\textbf{Qualitative comparisons with SDS-based methods.} Each column takes a unique text prompt of (a)\textit{``a yellow sports car with red wheel and tinted window''}, (b)\textit{``a white SUV with a blue police light on top of it''}, (c)\textit{``the chair has a red seat and yellow legs''}, (d)\textit{``a black office chair with a blue seat''}, (e)\textit{``a red bicycle with yellow pedals''}, and (f)\textit{``a green and white dirt bike''}.}
    \label{fig:sds_compare}
    %\vspace{-5mm}
\end{figure}

%% file: table/inference_time.tex
\begin{table}[t]
    \centering
    \caption{\textbf{Runtime comparisons for diffusion/GAN-based generative models.} We compare the inference time reported in~\cite{wei2023taps3d, lin2023magic3d}. For TAPS3D and TPA3D, we calculate the average time by generating 1000 samples with different text prompts.}
    %\vspace{-3mm}
    \resizebox{0.8\linewidth}{!}{
    \setlength{\tabcolsep}{12pt}
    \begin{tabular}{l|ccc}
    \toprule
    \multicolumn{1}{l|}{Method} & Device & Output & Time \\ \midrule 
    DreamFusion~\cite{poole2022dreamfusion} & TPUv4 machine & Rendering & 90 min \\
    Magic3D~\cite{lin2023magic3d}  & NVIDIA A100 x8 & Rendering & 40 min  \\
    TITG3SG~\cite{liu2022towards} & Telsa V100-32G & Voxel & 2.21 sec  \\
    TAPS3D~\cite{wei2023taps3d} & Telsa V100-32G & Rendering & 0.05 sec  \\
    TAPS3D~\cite{wei2023taps3d} & Telsa V100-32G & Mesh & 1.03 sec  \\
    \midrule
    Ours (TPA3D) & Telsa V100-32G & Rendering & 0.09 sec \\
    Ours (TPA3D )& Telsa V100-32G & Mesh & 2.87 sec  \\
    \bottomrule
    \end{tabular}}
    %\vspace{-2mm}
    \label{table:inference_time}
\end{table}

%% file: sec/6_conclusion.tex
\section{Conclusion}
\label{sec:conclusion}
%\subsubsection{Conclusion.}
In this paper, we proposed TPA3D, a GAN-based deep learning framework for fast text-guided 3D object generation. With only access to 3D shape data and their rendered 2D images, we utilized a pre-trained image captioning model and text encoder to generate detailed pseudo captions from the above visual data as the text condition. By observing the text condition, our TPA3D is able to extract geometry and texture triplane features for generating textured 3D meshes. Taking the sentence feature of the text description as input, the sentence-level generator of our TPA3D derives sentence-level triplane features. To enforce fine-grained details from the word-level descriptions, the introduced TPA block further performs word-level refinement during generation. From the experiments, we demonstrate the ability of TPA3D in matching the generated textured mesh to the detailed description while retaining sufficient fidelity.

% \subsubsection{Limitations.} Although our proposed TPA3D performs fast text-guided 3D generation with satisfactory alignment with text inputs, the retrieved text condition of TPA3D relies on pre-trained image captioning models and text encoders. While this suggests that our model does not limit the use of particular pre-trained vision-language models or text encoders, the quality of the resulting text embedding outputs would inevitably affect the learning and generation performances. 

\section*{Acknowledgement}
This work is supported in part by the National Science and Technology Council via grant NSTC 112-2634-F-002-007 and NSTC 113-2640-E-002-003, and the Center of Data Intelligence: Technologies, Applications, and Systems, National Taiwan University (grant nos.113L900902, from the Featured Areas Research Center Program within the framework of the Higher Education Sprout Project by the Ministry of Education (MOE) of Taiwan). We also thank the National Center for High-performance Computing (NCHC) for providing computational and storage resources.

%% file: sec/X_supp.tex
\clearpage

\centerline{\Large{\textbf{Appendix}}}

\appendix
\section{Implementation Details}
\label{sec:implementaion_details}
\subsection{Model Configuration}
In our TPA3D, we set $N = 6$ layers in the configuration of our sentence-level triplane generators $G$. As for embedding text inputs, we employ the pre-trained CLIP ViT-B/32~\cite{radford2021learning} as our encoder, keeping its weight frozen during training. In all experiments, we employ 8 NVIDIA V100 GPUs with a batch size of 32 to train both the generator and discriminator, completing the total training duration in approximately three days. 
Specifically, we use the Adam~\cite{kingma2014adam} optimizer with \(\beta = 0.9\), setting the learning rate to 0.001 for the generator and 0.002 for the discriminator. Finally, the whole model is implemented with the PyTorch~\cite{paszke2019pytorch} framework, and we render and visualize the generated 3D objects using Blender~\cite{blender}.

\subsection{More Details about Datasets}
In Sec.~\textcolor{red}{5.1}, we detail the preparation for datasets of ShapeNet~\cite{chang2015shapenet} and OmniObject3D~\cite{wu2023omniobject3d}. Specifically, we render 179,928, 162,672, 37,700, 19,200, and 9900 images with a resolution of 1024x1024 for \textit{Car, Chair, Motorbike,} \textit{Vehicle}, \textit{Accessory}, respectively. And following the setting provided by GET3D~\cite{gao2022get3d}, we split the training, validation, and testing set with the ratio of 7:1:2. Note that we also run the official implementation of TAPS3D~\cite{wei2023taps3d} and GET3D with the same data split for comparison.

\input{figure/supp/get3d}
\subsection{More Details about GET3D}
In Sec.~\textcolor{red}{3} of our main paper, we provide a brief review of GET3D~\cite{gao2022get3d}. Here, we detail the generator architecture of GET3D in Fig.~\ref{fig:get3d}. With modulated convolution layers~\cite{karras2020analyzing} (denoted as Mod Conv) conditioned on latent vectors $\mathbf{w}_{\text{geo}}$ and $\mathbf{w}_{\text{tex}}$, the geometry triplane $c^g_i$ and texture triplane $c^t_i$ are generated from each layer $i$.  By bilinear upsampling, the triplanes can be added together with the same resolution, and the final triplanes $f_{N}^{g}=\sum_{i=1}^N(c^g_i)$ and $f_{N}^{t}=\sum_{i=1}^N(c^t_i)$ are used for generating the output 3D textured mesh.

\input{figure/supp/psuedo_caption_pair}
\subsection{More Details of Pseudo Captioning}
As for the production of pseudo captions, we generate them using InstructBLIP~\cite{dai2023instructblip} with NVIDIA V100 GPUs beforehand to reduce the required computational resources during training. To elaborate, we load a pre-trained InstructBLIP (Vicuna-7B) and specify the input prompt as \textit{``In the image, the background is black. Describe the design and appearance of the \{category\} in detail.''} Subsequently, we refine the generated pseudo captions for each rendered image by filtering out redundant information. This involves eliminating background-related phrases like \textit{``in the black background''} and \textit{``with a black background''} to maintain focus on the geometry and texture of the 3D object. Additionally, distracting phrases such as \textit{``This is a 3D model of''} or \textit{``This is a 3D rendering of''} are removed to ensure a more precise and relevant description. In Figure~\ref{fig:pseudo_caption}, we sample several pseudo caption pairs with rendered RGB images of 3D objects from ShapeNet~\cite{chang2015shapenet} and OmniObject3D~\cite{wu2023omniobject3d}.

\input{table/ablation_components}
\input{figure/supp/TPA_removal}
\section{Ablation Study}
Since the proposed TPA module serves as the major contribution to our TPA3D, we conduct an ablation study on TPA blocks to quantitatively assess the impact of this module. In addition, we also conduct an ablation study on the training losses we introduced to analyze how training objectives influence the training process. All results are presented in Table~\ref{table:ablation_components} and Table~\ref{table:ablation_all}. Note that the baseline model (i.e., w/o TPA) removes both TPA$_{\text{geo}}$ and TPA$_{\text{tex}}$ and generates the final triplanes only using sentence-level triplanes conditioned on sentence features.

\subsection{Components in TPA blocks.}
\label{sec:ablation_components}
To verify the function of each component in TPA blocks as our claim, we singly remove cross-plane attention or cross-word attention in TPA blocks (as shown in Figure~\ref{fig:TPA_ablation}). From Table~\ref{table:ablation_components}, we can see that the removal of cross-plane attention leads to a significant decrease in FID. This result indicates that without the enhancement of plane features before cross-word attention, the visual quality severely decreases due to incomplete spatial information. On the other hand, the removal of cross-word attention brings a large drop in CLIP-R precision. This result verifies that cross-word attention promotes the correspondence between the detailed input text and the generated 3D object.

\input{table/ablation_all}
\subsection{TPA$_{\text{geo}}$ and TPA$_{\text{tex}}$.}
\label{sec:ablation_branch}
In order to generate textured meshes while achieving disentanglement of geometry and texture, TPA3D utilizes distinct branches to generate the final geometry and texture triplanes. To assess the efficacy of TPA blocks in these two branches, we symmetrically remove TPA$_{\text{geo}}$ and TPA$_{\text{tex}}$ and compare the results with the full architecture and baseline model without TPA blocks. The outcomes, presented in Table~\ref{table:ablation_branch}, reveal a slight decline in FID and CLIP R-precision when either geometry TPA blocks or texture TPA blocks are omitted. Compared to the baseline model without TPA, the additional TPA blocks on either branch can still enhance visual quality and text-shape consistency.

\subsection{Block Numbers of TPA.}
\label{sec:ablation_number}
In Table~\ref{table:ablation_number}, we compare models with varying numbers of TPA blocks. The original TPA3D applies six TPA blocks across all output resolutions of the sentence-level triplane generator $G$. To explore the impact of reducing the number of TPA${_{\text{geo}}}$ and TPA$_{\text{tex}}$, we create a variant with TPA blocks only applied in the first three resolutions of both the geometry and texture branches. From the findings shown in Table~\ref{table:ablation_number}, it is evident that a reduction in the number of TPA blocks corresponds to a decrease in both FID and CLIP R-precision metrics.

\subsection{Training Objectives.}
\label{sec:ablation_objective}
As noted in~\cite{wei2023taps3d}, training conditional GET3D~\cite{gao2022get3d} from scratch often results in collapsing shapes during training. To enable end-to-end one-stage training, we introduce additional $\mathcal{L}_{\text{mis}}$ in conjunction with $\mathcal{L}_{\text{clip}}$ to improve training stability and text-shape consistency. To investigate the impact of different objectives, we conduct an ablation study on these two training losses, $\mathcal{L}_{\text{mis}}$ and $\mathcal{L}_{\text{clip}}$. The results, as presented in Table~\ref{table:ablation_objective}, reveal that the absence of either $\mathcal{L}_{\text{mis}}$ or $\mathcal{L}_{\text{clip}}$  leads to degraded performance in both FID and CLIP R-precision. This further validates our inclusion of additional negative pairs with mismatched text conditions in $\mathcal{L}_{\text{mis}}$ significantly benefits the discriminator and enhances the stability of training, resulting in improvement in FID and CLIP R-precision.

\input{figure/supp/more_color_class}
\section{More Qualitative Results}
In this section, we present more qualitative results in terms of styles, subclasses, and more detailed text prompts to showcase TPA3D's ability to produce 3D shapes that closely align with the given text prompt. 

\subsection{Disentanglement of Geometry and Texture} In Figure~\ref{fig:more}, our TPA3D demonstrates its ability to disentangle geometry and texture information, resulting in shapes and textures that match the provided text prompt. By fixing the random seed and subclass in each row, the geometry of the generated shapes remains nearly unchanged, showcasing precise modifications to the texture. Similarly, each column illustrates that consistent color information can be applied to different geometries, even when different subclasses are specified in the text prompt. Additionally, our TPA3D can interpret rarer colors such as \textit{``scarlet''}, \textit{``light green''}, \textit{``aqua''}, and \textit{``indigo''}, generating textures that visually align with the given commands. In summary, our proposed TPA blocks excel in conducting word-level refinement for both texture and geometry, enabling precise generation to match the details specified in the given text prompt.

\input{figure/supp/interpolation}
\subsection{Interpolation of Geometry and Texture}
In addition to disentanglement, our TPA3D also keeps the continuity of the latent space for generating 3D shapes and textures. In Figure~\ref{fig:interpolation}, we use the same random noises $\mathbf{z}_{\text{geo}}$ and $\mathbf{z}_{\text{tex}}$ for each row, and then generate the shapes and textures with the interpolation on latent vectors $\mathbf{w}_{\text{geo}}$ and $\mathbf{w}_{\text{tex}}$ for different text prompts. We can see the continuous shapes and colors match the interpolation of text prompts, which demonstrates the continuity of the latent space.

\input{figure/supp/more_compare}
\subsection{Generation for Detailed Description}We provide more qualitative comparisons with TAPS3D~\cite{wei2023taps3d} to show the ability to generate objects with detailed text descriptions in Figure~\ref{fig:more_compare}. The results presented in Figure~\ref{fig:more_compare} confirm that TPA3D effectively leverages word-level information to refine the textured shape, ensuring alignment with the specific details mentioned in the textual description. Considering the generation of a military-style SUV as an example, TPA3D not only accurately captures the geometry of the \textit{``SUV''} but also successfully retrieves detailed requirements such as \textit{``military-style''} and \textit{``camouflage paint''}. With word-level refinement provided by TPA blocks, our TPA3D also excels in specifying modifications to particular elements such as \textit{``white roof''}, \textit{``striped seat and backrest''} or \textit{``with a band around it''}. In contrast, TAPS3D can only generate textured shapes at a more generalized class and color level. For a comprehensive evaluation of the 3D shape from various perspectives, a \textit{.mp4} file (\textit{detailed\_text\_prompt.mp4}) is provided for comparative analysis with TAPS3D.

\input{figure/supp/more_manipulation}
\subsection{Controllable Manipulation}In Figure~\ref{fig:more_manipulation}, we present additional manipulation experiments, offering users the ability to incrementally adjust the generated shape according to their specific stylistic and geometric preferences by fixing random seeds. For instance, after generating a \textit{``white motorbike''}, users can add \textit{``purple''} decorations to the same bike or even append \textit{``a yellow headlight''} to the front of the motorbike. Likewise, users can incorporate geometry details such as \textit{``armrests''} to the original simplistic \textit{``red chair''}, maintaining nearly unchanged shapes and textures. This capability to produce 3D shapes rapidly and accommodate incremental text prompt requirements without significantly altering the initial shape enhances the utility of TPA3D as an interactive 3D asset creation tool, providing users with quick customization and control over the final output.

\input{figure/supp/multi}
\input{table/ablation_multi}
\subsection{Multi-class Generation}
To explore the potential of our TPA3D in the multi-class generation, we trained the model on a combined dataset comprising \textit{Car}, \textit{Chair}, and \textit{Motorbike}. We conducted both qualitative and quantitative evaluations to assess its performance. By comparing with the single-class version of TPA3D, the quantitative result in Table~\ref{table:multi} indicates that our TPA3D is capable of multi-class generation. Given the same model capacity as the single-class version, we anticipate a minor decrease in FID and CLIP R-precision scores, but the results remain satisfactory. Also, as depicted in Figure~\ref{fig:multi}, the multi-class adaptation of TPA3D yields 3D textured shapes closely matching the detailed text prompts in texture and geometry.

\section{Human Evaluation for Verifying Fidelity}
We conduct a human evaluation on 40 subjects to support our TPA3D. We use the same 100 captions for both our method and TAPS3D to generate 40, 40, and 20 objects for \textit{Car}, \textit{Chair}, and \textit{Motorbike} in ShapeNet, respectively. Out of 4000 total responses, 2713 (67.8\%) favor our method. Besides, our method is preferred in 73 out of 100 examples. Lastly, all 40 subjects indicate that our method generates higher fidelity shapes. These results demonstrate our method's superior performance in fidelity.
It's worth noting that, quantitative evaluation has been provided in our main paper, which confirms that our method achieved better image fidelity than TAPS3D in terms of FID by a significant margin of 7.2 on average on ShapeNet, as shown in Table \textcolor{red}{1} in our main paper.

\input{figure/supp/simple_captions}
\section{TPA Effectiveness for Simple or Complex Captions}
As our major contribution, TPA blocks are designed to enhance triplane features for capturing fine-grained information from the input text, especially when the text descriptions are complex and with details. To verify this claim, we test our model and TAPS3D~\cite{wei2023taps3d} with only simple captions (only \textit{color} + \textit{class}) from \textit{Car} and \textit{Chair} in ShapeNet~\cite{chang2015shapenet}. As Shown in Figure~\ref{fig:simple_caption}, we can observe that the performance gap between our model and TAPS3D increases when we shift from simple captions to complex captions. For FID, the performance gap increases from 14.1\% to 29.9\% for \textit{Car} and from 3.7\% to 14.8\% for \textit{Chair}. As for CLIP R-precision, the performance gap increases from 500\% to 545\% for \textit{Car} and from 195\% to 413\% for \textit{Chair}. This result proves that our TPA blocks have better capability to deal with text prompts with detailed information.

\section{Limitations}
Although our proposed TPA3D performs fast text-guided 3D generation with satisfactory alignment with text inputs, the retrieved text condition of TPA3D relies on pre-trained image captioning models and text encoders. While this suggests that our model does not limit the use of particular pre-trained vision-language models or text encoders, the quality of the resulting text embedding outputs would inevitably affect the learning and generation performances. 

%% file: figure/supp/get3d.tex
\begin{figure}
    \centering
    \includegraphics[width=1\linewidth]{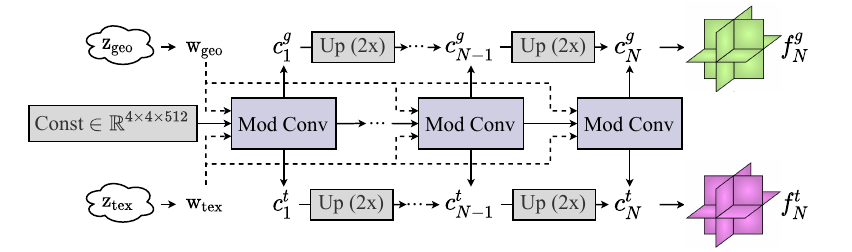}
    \caption{\textbf{Detailed architecture of GET3D~\cite{gao2022get3d}.} Referenced from the original paper of GET3D, the design of GET3D processed the produced triplane features, where \textit{Mod Conv} denotes the generator layers in GET3D, and $c^g_i$ and $c^t_i$ represents the $i$-th geometry and texture triplane, respectively. Note that the upsampling algorithm is bilinear upsampling.}
    \label{fig:get3d}
\end{figure}

%% file: figure/supp/psuedo_caption_pair.tex
\begin{figure}[tp]
    \centering
    \includegraphics[width=1\linewidth]{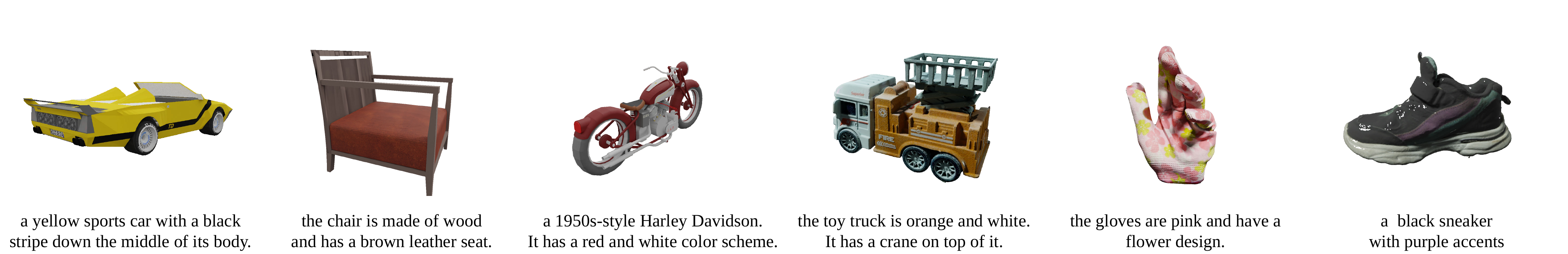}
    \caption{\textbf{Rendered images of 3D objects and their corresponding pseudo captions predicted by InstructBLIP~\cite{dai2023instructblip}.}} 
    \label{fig:pseudo_caption}
\end{figure}

%% file: table/ablation_components.tex
\begin{table}[tp]
    \centering
    \caption{\textbf{Ablation studies to the components in TPA blocks.} Note that FID and CLIP R-precision@5 are reported.}
    \label{table:ablation_components}
    \resizebox{1\textwidth}{!}{
        \setlength{\tabcolsep}{10pt}
        \begin{tabular}{l|ccc|ccc}
            \toprule
            \multirow{2}{*}{Method} & \multicolumn{3}{c|}{FID $\downarrow$} & \multicolumn{3}{c}{CLIP R-precision $\uparrow$} \\ 
            \cmidrule(lr){2-4} \cmidrule(lr){5-7}
            & Car & Chair & Motorbike & Car & Chair & Motorbike \\
            \midrule
            Ours & \textbf{18.5} & \textbf{38.1} & \textbf{77.7} & \textbf{80.94} & \textbf{38.58} & \textbf{24.76} \\
            w/o cross-word attn & 20.4 & 42.4 & 78.7 & 70.94 & 30.96 & 21.36 \\
            w/o cross-plane attn & 30.0 & 47.7 & 80.9 & 79.73 & 35.49 & 23.04 \\
            w/o TPA & 32.6 & 51.1 & 82.5 & 68.72 & 26.90 & 20.82 \\
            \bottomrule
        \end{tabular}
    }
\end{table}

%% file: figure/supp/TPA_removal.tex
\begin{figure}[tp]
    \centering
    \begin{subfigure}{0.49\linewidth}
        \includegraphics[width=1\linewidth]{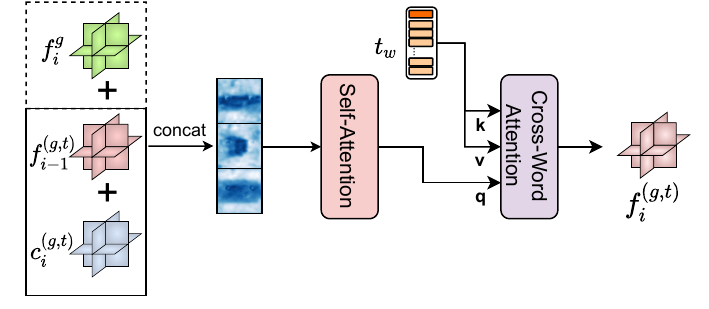}
        \caption{w/o cross-plane attention}
    \end{subfigure}
    \begin{subfigure}{0.49\linewidth}
        \includegraphics[width=1\linewidth]{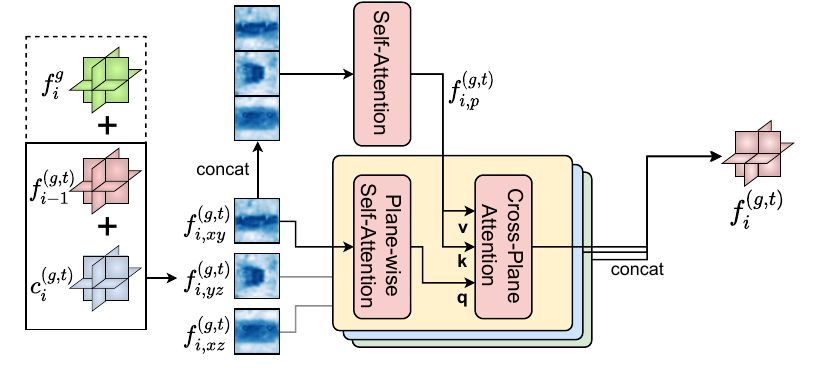}
        \caption{w/o cross-word attention}
    \end{subfigure}
    \caption{\textbf{The architectures for the ablation study in Sec.~\ref{sec:ablation_components}.} (a) Without cross-plane attention, the word features might be attended to the region with incomplete spatial information, which leads to a lower visual quality. (b) Without cross-word attention, triplanes lack detailed information in the description and only contain global information from sentence features.}
    \label{fig:TPA_ablation}
\end{figure}

%% file: table/ablation_all.tex
\begin{table}
    \centering
    \begin{subtable}[c]{0.49\textwidth}
        \centering
        \resizebox{0.8\textwidth}{!}{
            \setlength{\tabcolsep}{8pt}
            \begin{tabular}{l|cc}
                \toprule
                \multicolumn{1}{l}{} & FID $\downarrow$ & CLIP-R $\uparrow$ \\ \midrule
                Ours & \textbf{18.5} & \textbf{80.94}  \\
                w/o TPA$_{\text{tex}}$ & 18.6 &  76.85 \\
                w/o TPA$_{\text{geo}}$ & 20.4 & 79.46  \\
                w/o TPA  & 32.6 & 68.72 \\
                \bottomrule
            \end{tabular}
        }
        \caption{\textbf{TPA$_{\text{geo}}$ and TPA$_{\text{tex}}$ (Sec.~\ref{sec:ablation_branch})}}
        \label{table:ablation_branch}
    \end{subtable}
    \begin{subtable}[c]{0.49\textwidth}
        \centering
        \resizebox{1\textwidth}{!}{
            \setlength{\tabcolsep}{8pt}
            \begin{tabular}{l|cc}
                \toprule
                \multicolumn{1}{l}{} & FID $\downarrow$ & CLIP-R $\uparrow$ \\ \midrule
                Ours (full) & \textbf{18.5} & \textbf{80.94}  \\
                w/ TPA x 3 (half)& 20.1 & 78.07 \\
                w/ TPA x 0 & 32.6 & 68.72  \\
                \bottomrule
            \end{tabular}
        }
        \caption{\textbf{Block numbers of TPA (Sec.~\ref{sec:ablation_number})}}
        \label{table:ablation_number}
    \end{subtable}
    \begin{subtable}[c]{0.49\textwidth}
        \centering
        \resizebox{0.8\textwidth}{!}{
            \setlength{\tabcolsep}{8pt}
            \begin{tabular}{l|cc}
                \toprule
                \multicolumn{1}{l}{} & FID $\downarrow$ & CLIP-R $\uparrow$  \\ \midrule
                Ours & \textbf{18.5} & \textbf{80.94}  \\
                w/o $\mathcal{L}_{\text{clip}}$ & 56.9 & 56.62 \\
                w/o $\mathcal{L}_{\text{mis}}$& 68.6 & 68.19 \\
                \bottomrule
            \end{tabular}
        }
        \caption{\textbf{Training objectives (Sec.~\ref{sec:ablation_objective})}}
        \label{table:ablation_objective}
    \end{subtable}
    \caption{\textbf{Other ablation studies of our TPA3D on \textit{Car} in ShapeNet.} Note that FID and CLIP R-precision@5 are reported.}
    \label{table:ablation_all}
\end{table}

%% file: figure/supp/more_color_class.tex
\begin{figure*}[tp]
\setlength{\linewidth}{\textwidth}
\setlength{\hsize}{\textwidth}
    \centering
    \includegraphics[width=1\linewidth]{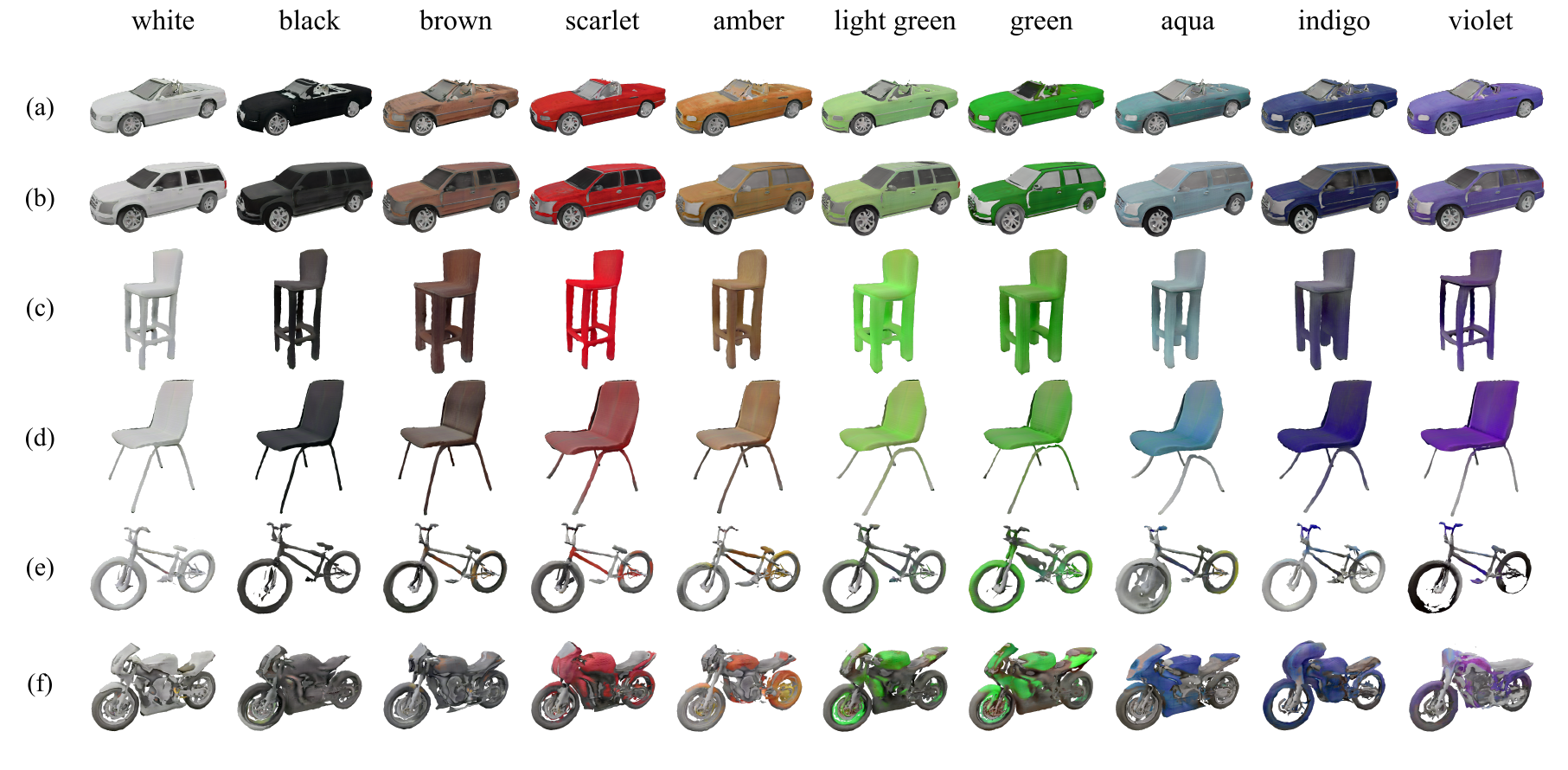}
    \caption{\textbf{More text-guided 3D generation results of TPA3D.} We formulate the input prompts as ``a \{color\} \{object\}'' with various colors and sub-classes for generation.
    Specifically, each column stands for a different color, while each row stands for a unique sub-class: (a) 
    \textit{``convertible''} (b) \textit{``SUV''} (c) \textit{``stool''} (d) \textit{``plastic chair''} (e) \textit{``bicycle''} (f) \textit{``sport bike''}}
    \label{fig:more}
\end{figure*}

%% file: figure/supp/interpolation.tex
\begin{figure}[tp]
    \centering
    \includegraphics[width=0.85\linewidth]{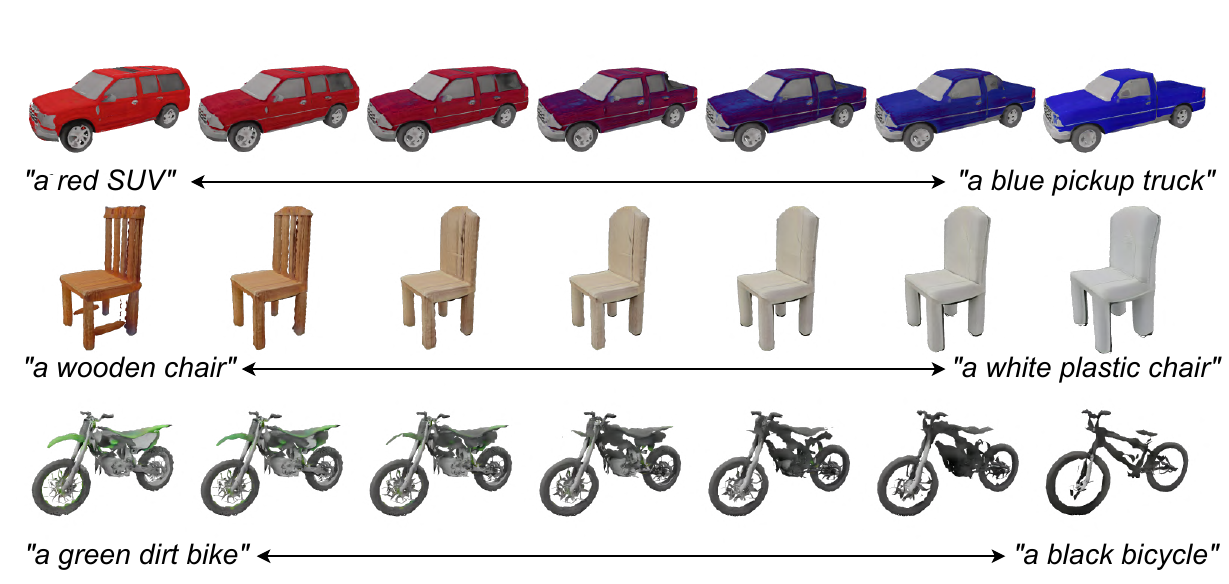}
    \caption{\textbf{The interpolation results of our TPA3D.} In each row, we use the same random noises $\mathbf{z}_{\text{geo}}$ and $\mathbf{z}_{\text{tex}}$, and perform interpolation on latent vectors $\mathbf{w}_{\text{geo}}$ and $\mathbf{w}_{\text{tex}}$ for different text inputs.}
    \label{fig:interpolation}
\end{figure}

%% file: figure/supp/more_compare.tex
\begin{figure}[tp]
    \centering
    \begin{subfigure}{1\linewidth}
        \includegraphics[width=1\linewidth]{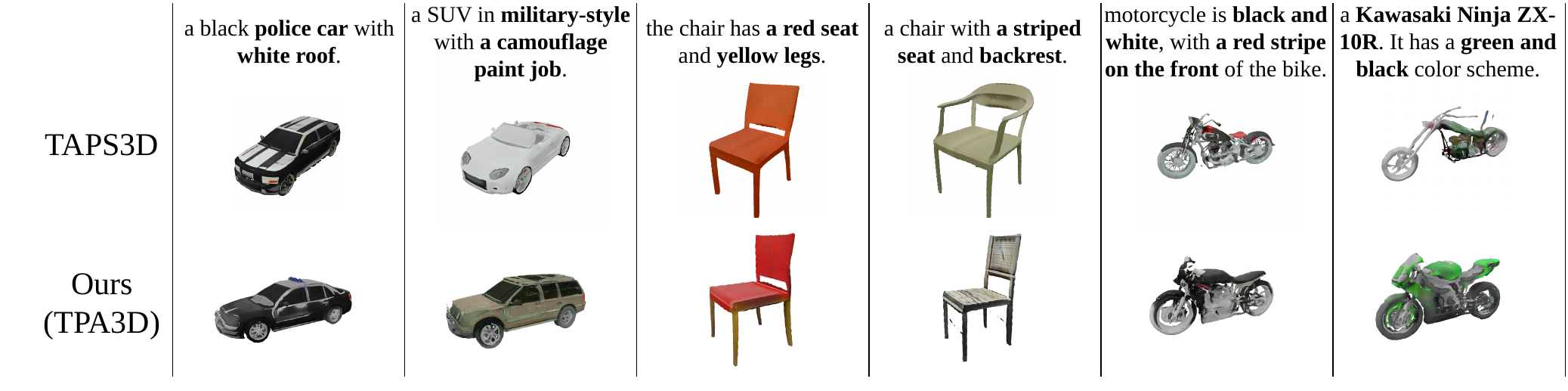}
        \caption{ShapeNet~\cite{chang2015shapenet}}
    \end{subfigure}
    \begin{subfigure}{1\linewidth}
        \includegraphics[width=1\linewidth]{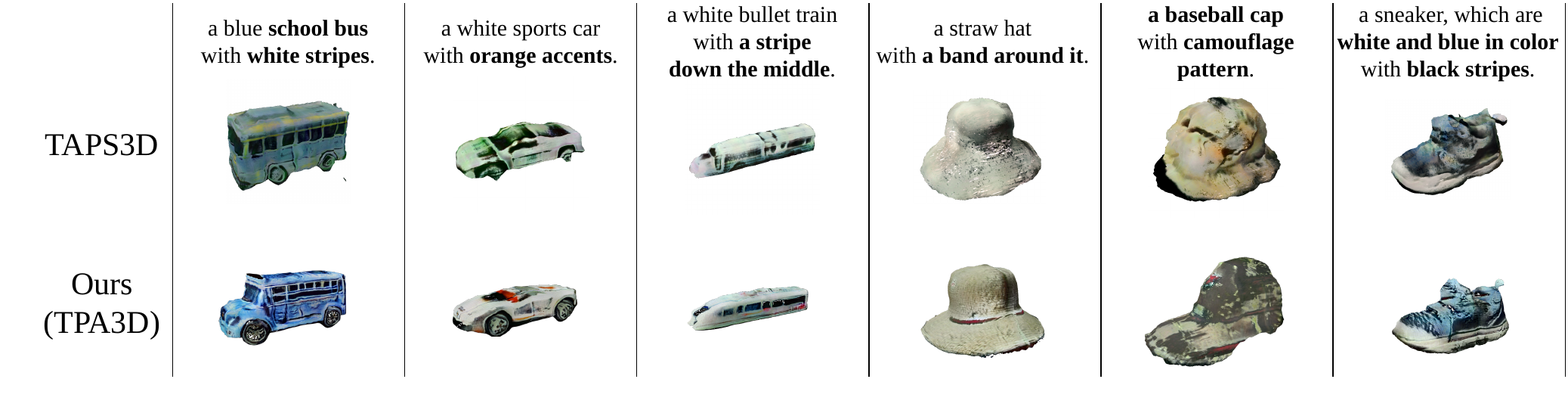}
        \caption{OmniObject3D~\cite{wu2023omniobject3d}}
    \end{subfigure}
    \caption{\textbf{More qualitative comparisons with TAPS3D~\cite{wei2023taps3d} on (a) ShapeNet and (b) OmniObject3D.} Given detailed textual descriptions, our TPA3D generates accurate shapes aligned to the texts, while TAPS3D only realizes general classes and simple colors.}
    \label{fig:more_compare}
\end{figure}

%% file: figure/supp/more_manipulation.tex
\begin{figure}[tp]
    \centering
    \includegraphics[width=0.8\linewidth]{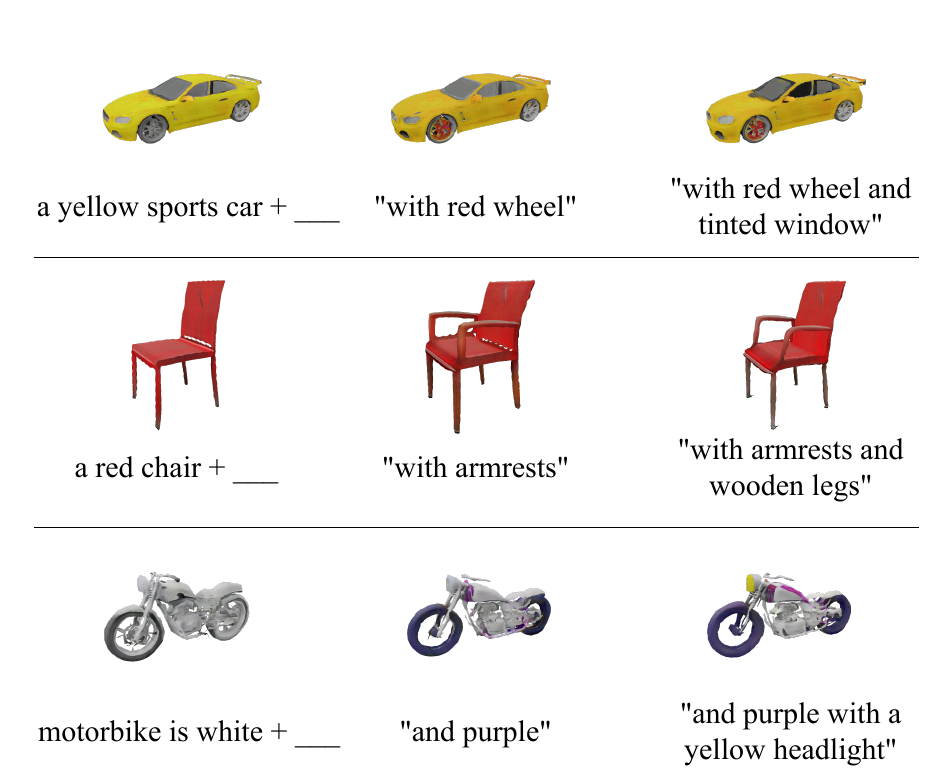}
    \caption{\textbf{More manipulation examples of adding different detailed text descriptions.} Each row shows an example of manipulating the left-most object with detailed descriptions. With the same random seed for sampling $\mathbf{z}_{\text{geo}}$ and $\mathbf{z}_{\text{tex}}$, two distinct results are shown along with the original one in each row.}
    \label{fig:more_manipulation}
\end{figure}

%% file: figure/supp/multi.tex
\begin{figure}[tp]
    \centering
    \includegraphics[width=1\linewidth]{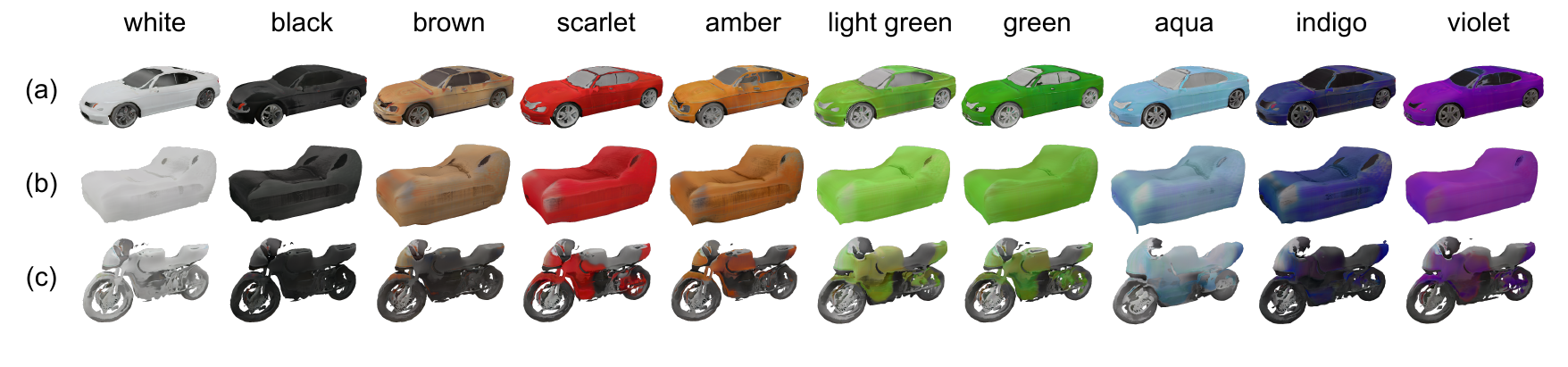}
    \caption{\textbf{Qualitative results of multi-class 3D generation.}  We formulate the input prompts as ``a \{color\} \{object\}'' with various colors and sub-classes for multi-class generation.
    Specifically, each column stands for a different color, while each row stands for a unique sub-class: (a) 
    \textit{``sports car''} (b) \textit{``sofa''} (c) \textit{``sport bike''}}
    \label{fig:multi}
\end{figure}

%% file: table/ablation_multi.tex
\begin{table}[tp]
    \centering
    \caption{\textbf{Comparisons between single-class and multi-class TPA3D.} For multi-class generation, we train our TPA3D using data of \textit{Car}, \textit{Chair} and \textit{Motorbike}. Following the single-class evaluation protocol, we assess this multi-class TPA3D for each class by randomly sampling pseudo captions from the respective test set to generate rendered images, with both FID and CLIP R-precision@5 as the metrics. With an equivalent model capacity, the outcomes exhibit a slight degradation but remain satisfactory.}
    \resizebox{0.8\textwidth}{!}{
        \setlength{\tabcolsep}{10pt}
        \begin{tabular}{c|c|cc}
            \toprule
            Class & Method & FID($\downarrow$) & CLIP R-precision($\uparrow$) \\ \midrule \midrule
            \multirow{2}{*}{Car} & Ours (single-class) & 18.50 & 80.94 \\
            {} & Ours (multi-class) & 20.77 & 68.79 \\
            \midrule
            \multirow{2}{*}{Chair} & Ours (single-class) & 38.11 & 38.58 \\
            {} & Ours (multi-class) & 47.54 & 20.80 \\
            \midrule
            \multirow{2}{*}{Motorbike} & Ours (single-class) & 77.69 & 24.76 \\
            {} & Ours (multi-class) & 74.79 & 19.00 \\
            \bottomrule
        \end{tabular}
    }
    \label{table:multi}
\end{table}

%% file: figure/supp/simple_captions.tex
\begin{figure}[tp]
    \centering
    \includegraphics[width=1\linewidth]{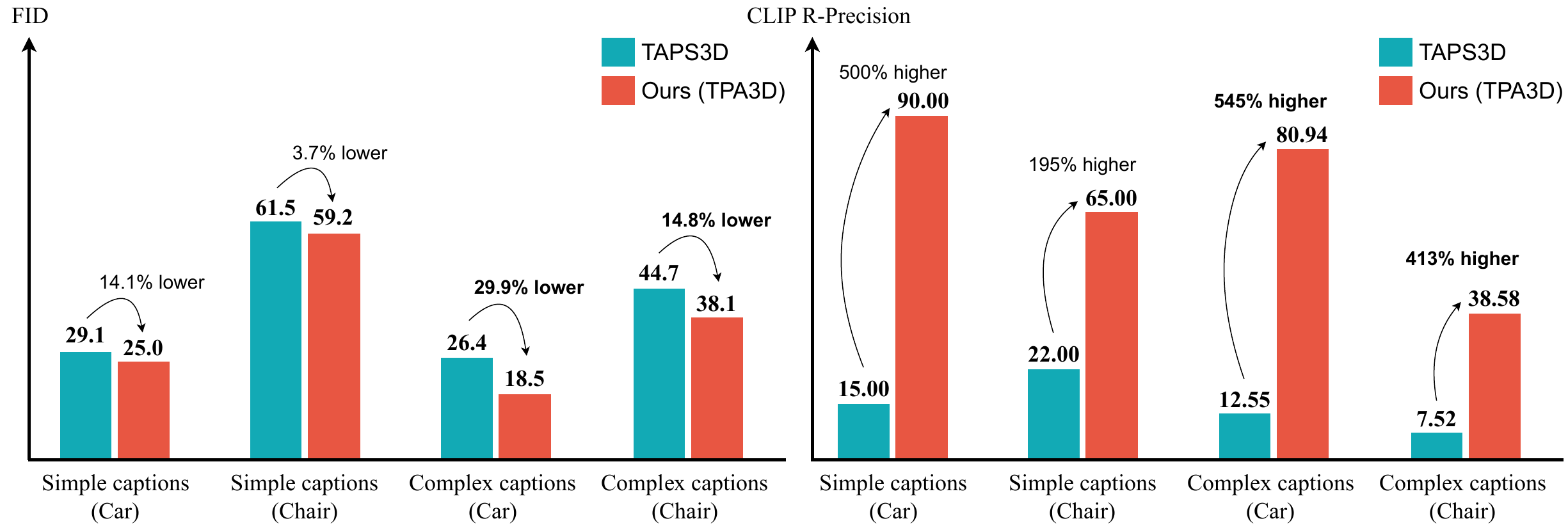}
    \caption{\textbf{Performance comparison with TAPS3D~\cite{wei2023taps3d} using simple and complex captions in terms of FID and CLIP R-precision.} Different from complex captions containing fine-grained descriptions, simple captions are only composed of \textit{color} and \textit{class}. The performance gap is larger when using complex captions, which validates the effectiveness of our TPA blocks in dealing with detailed text prompts.}
    \label{fig:simple_caption}
\end{figure}